\documentclass[runningheads]{llncs}
\usepackage{graphicx}
\usepackage[pagebackref,breaklinks,colorlinks]{hyperref}

\usepackage{tikz}
\usepackage{comment}
\usepackage{amsmath,amssymb} 
\usepackage{color}
\usepackage{booktabs}
\usepackage{diagbox}
\usepackage{tabu}
\usepackage{rotating}
\usepackage{array,multirow}
\usepackage{float}
\usepackage{enumitem}
\usepackage{wrapfig}
\usepackage{pifont}
\usepackage{algorithmic}
\usepackage{algorithm}

\def\eg{\emph{e.g}} 
\def\ie{\emph{i.e}} 
 
\def\etc{\emph{etc}}

\def\etal{\emph{et al}}

\newcommand\myeq{\mkern1.5mu{=}\mkern1.5mu}

\usepackage[accsupp]{axessibility}  

\begin{document}
\pagestyle{headings}
\mainmatter
\def\ECCVSubNumber{4408}  

\title{Invariant Feature Learning for Generalized Long-Tailed Classification} 


\titlerunning{Generalized Long-Tailed Classification}

\author{
    Kaihua Tang\inst{1},
    Mingyuan Tao\inst{2},
    Jiaxin Qi\inst{1}\thanks{Corresponding author.},
    Zhenguang Liu\inst{3}, 
    Hanwang Zhang\inst{1}
}
\authorrunning{K. Tang et al.}
%
\institute{
    Nanyang Technological University, Singapore \and
    Damo Academy, Alibaba Group, Hangzhou, China \and
    Zhejiang University, Hangzhou, China \\
    \email{kaihua.tang@ntu.edu.sg}; \email{juchen.tmy@alibaba-inc.com}; \email{jiaxin003@e.ntu.edu.sg} \\
    \email{liuzhenguang2008@gmail.com}; \email{hanwangzhang@ntu.edu.sg}
}


\maketitle

\begin{abstract}
Existing long-tailed classification (LT) methods only focus on tackling the \textbf{class-wise imbalance} that head classes have more samples than tail classes, but overlook the \textbf{attribute-wise imbalance}. In fact, even if the class is balanced, samples within each class may still be long-tailed due to the varying attributes. Note that the latter is fundamentally more ubiquitous and challenging than the former because attributes are not just implicit for most datasets, but also combinatorially complex, thus prohibitively expensive to be balanced. Therefore, we introduce a novel research problem: \textbf{Generalized Long-Tailed} classification (GLT), to jointly consider both kinds of imbalances. By ``generalized'', we mean that a GLT method should naturally solve the traditional LT, but not vice versa. Not surprisingly, we find that most class-wise LT methods degenerate in our proposed two benchmarks: ImageNet-GLT and MSCOCO-GLT. We argue that it is because they over-emphasize the adjustment of class distribution while neglecting to learn attribute-invariant features. To this end, we propose an Invariant Feature Learning (IFL) method as the first strong baseline for GLT. IFL first discovers environments with divergent intra-class distributions from the imperfect predictions, and then learns invariant features across them. Promisingly, as an improved feature backbone, IFL boosts all the LT line-up: one/two-stage re-balance, augmentation, and ensemble. Codes and benchmarks are available on Github: \url{https://github.com/KaihuaTang/Generalized-Long-Tailed-Benchmarks.pytorch}
\end{abstract}

\section{Introduction}
\label{sec:1}

\begin{figure}[t]
   \begin{minipage}[b]{1.0\linewidth}
   \centerline{\includegraphics[width=125mm]{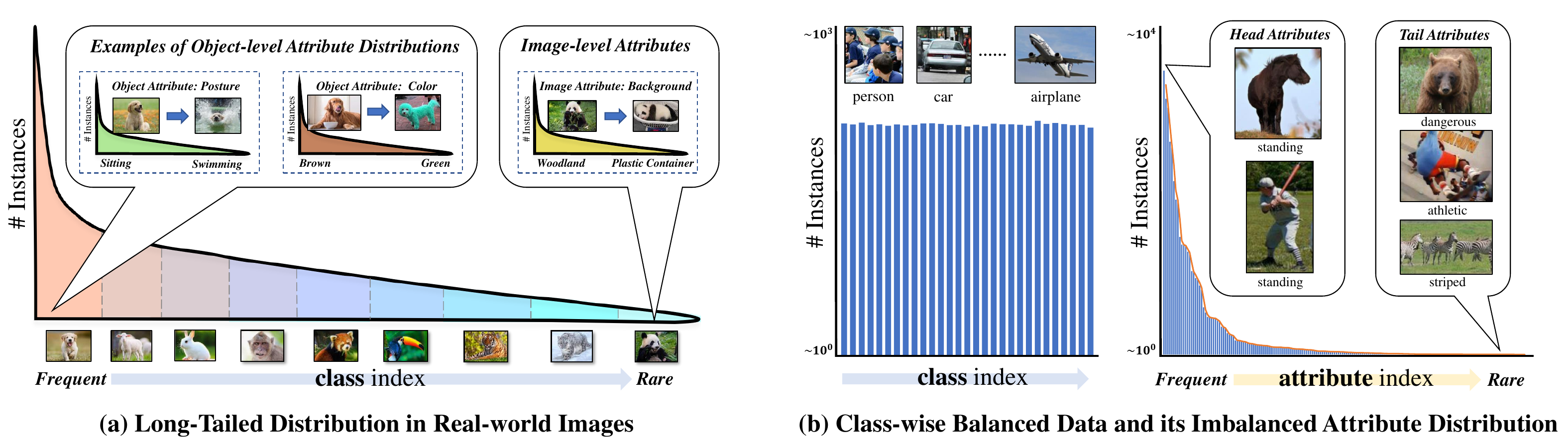}}
   \end{minipage}
   \caption{(a) The real-world long-tailed distribution is both class-wise and attribute-wise imbalanced; (b) even if we balance the class distribution of MSCOCO-Attribute~\cite{patterson2016coco}, the attributes are still long-tailed}
   \label{fig:1} 
\end{figure}

Long-Tailed classification (LT)~\cite{zhang2021deep} is inevitable in real-world training, as long-tailed distribution ubiquitously exists in data at scale~\cite{reed2001pareto,powers1998applications} and it is often prohibitively expensive to balance against such nature~\cite{reed2001pareto}. For example in Fig.~\ref{fig:1}~(a), a frequent class such as ``dog'' has significantly more samples than a rare one such as ``panda''. Prevailing LT methods are essentially based on adjusting the \textbf{class-wise} imbalance ratio: given a biased classifier that tends to classify tail as head, we curb the head confidence while lifting the tail confidence, so the resultant classifier is expected to be fair to both head and tail during inference~\cite{kang2019decoupling,zhou2019bbn,cao2019ldam,menon2020long,ren2020balanced}.

\begin{wrapfigure}{R}{70mm}
\includegraphics[width=70mm]{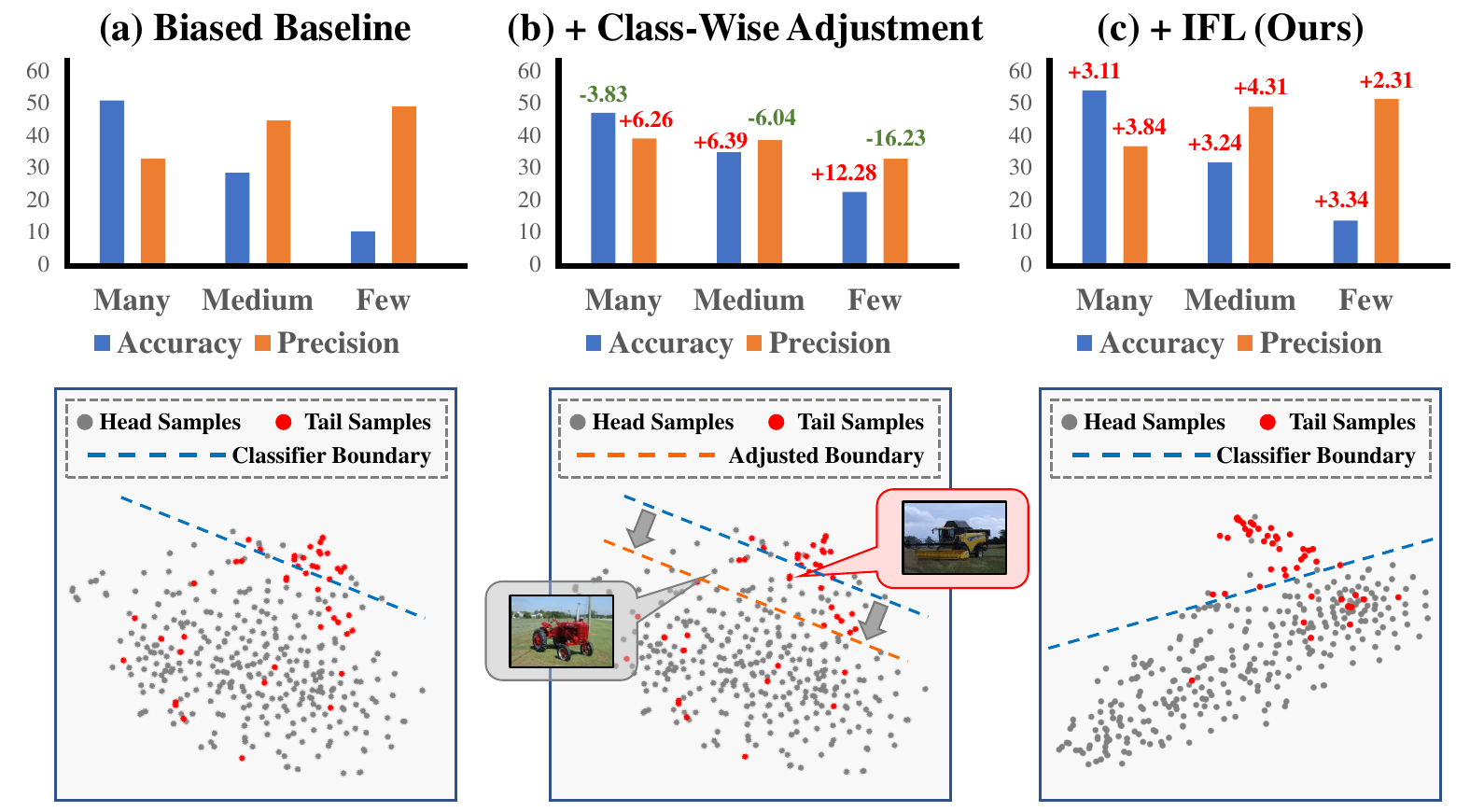}
\caption{In ImageNet-GLT, a typical LT method, (b) LWS~\cite{kang2019decoupling}, is playing a precision-accuracy trade-off with the (a) biased cross-entropy baseline, while the proposed (c) IFL improves both metrics at the same time. We follow \cite{liu2019large,kang2019decoupling} to stratify classes into Many, Medium, and Few by the class frequency. The t-SNE~\cite{hinton2002stochastic} of image features further illustrates that IFL features reduce the confusing region between ``tractor'' and ``harvester'' caused by the shared attribute ``field''}
\label{fig:2} 
\end{wrapfigure}

However, we'd like to point out that LT challenge cannot be simply characterized by class-wise imbalance. If we take a closer look at samples inside each class in Fig.~\ref{fig:1}~(a), we can find that attributes within each class are also long-tailed. This \textbf{attribute-wise} imbalance\footnote{In this paper, the attribute represents all the factors causing the intra-class variations, including object-level attributes (colors, textures, postures, \etc) and image-level attributes (lighting, contexts, \etc).} undermines the robustness of the classifier in two ways: 
\textbf{I)} it hurts the accuracy of images with tail attributes, \eg, members of minority groups in the human class are easier to be mis-classified than their majority counterparts~\cite{bbc_news_2021}, despite the fact that they both come from the same head class; 
\textbf{II)} it results in some attributes being mistakenly correlated to certain classes, \eg, images of a head class ``tractor'' are often captured on ``field'', so when a tail object like ``harvester'' is also captured on ``field'', its risk of being mis-classified as ``tractor'' is much higher, which is supported by our formulation Eq.~\eqref{eq.1} and visualization in Fig.~\ref{fig:2}~(b). Therefore, the \textbf{attribute-wise} imbalance further explains the cause of inconsistent performances within the same class and the existence of spurious correlations.

In fact, even if the class-wise imbalance is entirely eliminated like Fig.~\ref{fig:1}~(b), its attribute-wise imbalance still persists and hurts the generalization. Besides, strictly balancing attributes is not only prohibitive but also impossible due to the innumerable multi-label combinations of attributes, making the attribute-wise imbalance fundamentally different from the class-wise imbalance. To this end, we present a new task: \textbf{Generalized Long-Tailed} classification (GLT), to unify the challenges from both class-wise and attribute-wise imbalances. For rigorous and reproducible evaluations in the community, as detailed in Section~\ref{sec:3}, we introduce two benchmarks, ImageNet-GLT and MSCOCO-GLT, together with three protocols to evaluate the robustness of models against class-wise long tail, attribute-wise long tail and their joint effect.




Not surprisingly, we find that nearly all existing LT methods~\cite{zhu2022cross,menon2020long,kang2019decoupling,ren2020balanced} fail to tackle the attribute-wise imbalance in GLT (cf. Section~\ref{sec:5}). The reasons are two-fold: 
\textbf{I)} They rely on class-wise adjustment, which requires the access to the class statistical traits to re-balance~\cite{tan2020equalization,cao2019ldam,kang2019decoupling,zhou2019bbn}. Unfortunately, the attribute-wise traits are hidden in GLT, whose discovery \emph{per se} is a challenging open problem~\cite{li2021discover}. 
\textbf{II)} As illustrated in Fig.~\ref{fig:2}, the cross-entropy baseline (biased classifier) tends to predict tail samples as the head class with similar attributes, resulting in low precision on the head and low accuracy on the tail. So, the success of LT methods is mainly based on lifting the tail class boundary to welcome more samples to increase the tail accuracy. However, such adjustment is only playing with the precision-accuracy trade-off~\cite{zhu2021cross}, leaving the confused region of similar attributes unchanged in the feature space, while the proposed GLT requires algorithms to ignore those confusing attributes.




To this end, in Section~\ref{sec:4}, we introduce a framework called Invariant Feature Learning (IFL) to address the attribute-wise imbalance and serve as the first strong baseline for GLT. Our motivation is based on the reasonable assumption that: since the class feature is invariant to its attributes, \eg, a ``dog'' is always a dog regardless of its varying attributes, the variation of attributes is the main cause of lower prediction confidence within each class. Note that corrupted images are beyond the scope of this paper, as they don't have any valid labels. Therefore, we use the current classification confidence of each training sample as an imbalance indicator of attributes inside the class: the lower the confidence is, the rarer the attributes are. Then, we sample a new environment based on the reversed confidence. Together with the original one, we obtain two environments with diverse attribute distributions for each class. Finally, to remove the imbalance impact of attributes, we design a metric learning loss, extending the center loss~\cite{wen2016centerloss} to its Invariant Risk Minimization (IRM)~\cite{arjovsky2019invariant} version, which equips the model with the ability to learn class features that are invariant to attributes. 



As shown in Fig.~\ref{fig:2}~(c), IFL improves both precision and accuracy under GLT by reducing the confusing region of attributes in the feature space. Besides, IFL, as an improved feature backbone, can be seamlessly incorporated into other LT methods. In particular, we find that by only using sample augmentation such as MixUp~\cite{zhang2018mixup} or RandAug~\cite{cubuk2020randaugment}, IFL can surpass most of the LT methods. We also notice that the recent progress of LT methods~\cite{wang2020long,zhang2021test,zhu2021cross}, who claim to improve both head and tail, is indeed attributed to tackle the attribute-wise imbalance---they deserve to be more fairly evaluated by GLT.



Our contributions can be summarized as follows:
\begin{itemize}
\item We present a new challenging task: Generalized Long-Tailed classification (GLT), together with two benchmarks: ImageNet-GLT and MSCOCO-GLT. To solve GLT, one need to address both the conventional class-wise imbalance and the ever-overlooked attribute-wise imbalance.

\item We develop Invariant Feature Learning (IFL) as the first strong GLT baseline. Its effectiveness demonstrates that learning attribute-invariant features is a promising direction.

\item By extensive experiments, we show that IFL improves all the prevalent LT line-up on GLT benchmarks: one/two-stage re-balancing~\cite{kang2019decoupling,tang2020long,ren2020balanced,menon2020long,zhou2019bbn,cao2019ldam}, augmentation~\cite{zhang2018mixup,cubuk2020randaugment}, and ensemble~\cite{wang2020long,zhang2021test}.
\end{itemize}

\section{Related Work}
\label{sec:2}

\noindent\textbf{Long-Tailed Classification}~\cite{zhu2022cross,liu2019large,cui2019class,zhang2021deep} aims to improve the performance under class-wise balanced evaluation given the class-wise long-tailed training data. Previous methods can be categorized into three types: 1) one/two-stage re-balancing algorithms~\cite{kang2019decoupling,hu2020learning,tan2021equalization,li2020overcoming,tan2020equalization,wang2020devil,hong2021disentangling} apply statistical adjustment based on the explicit class distribution to correct the tail class bias; 2) data augmentation either independently augments all samples~\cite{yin2019feature,kim2020m2m,zhang2018mixup,cubuk2020randaugment} or transfer head information to the tail~\cite{liu2020deep,he2021distilling,li2021self,wang2017learning}; 3) ensemble~\cite{wang2020long,cai2021ace,xiang2020learning,zhang2021test} is recently explored as a strategy to improve head and tail categories at the same time. The conventional LT classification is essentially a special case of the proposed GLT, as solving GLT will naturally improve LT, but not vice versa.

\noindent\textbf{Domain Adaptation (DA) and Out-of-distribution Generalization (O-ODG)} are two other related tasks. DA~\cite{wilson2020survey,wang2018deep,zhao2019learning,yue2021transporting} seeks to transfer models from source domains to the target domain. The difference between DA and GLT is that we don't need a subset of samples from the target domain. Recent papers~\cite{jamal2020rethinking,zou2018unsupervised} also notice the intrinsic correlation between LT and DA. In OODG~\cite{zhao21robin,arjovsky2020out,krueger2021out,arjovsky2019invariant}, we desire a machine trained in one domain to work well in any domain. Recent studies~\cite{arjovsky2020out} show that it cannot be addressed without any assumption. Therefore, regarding the type of assumptions, OODG can be divided into DA~\cite{wang2018deep}, domain generalization~\cite{li2017deeper}, long-tailed classification~\cite{zhang2021deep}, zero-/few-shot learning~\cite{sung2018learning,wang2019few,wang2019survey}, and even adversarial robustness~\cite{carlini2019evaluating,chakraborty2018adversarial}. Our GLT can also be viewed as a special case of OODG that is more general than LT.

\noindent\textbf{Attribute-wise Imbalance} itself is also a long-standing research field in the name of hard example mining~\cite{lin2017focal,nam2020learning}, sub-population shift~\cite{santurkar2020breeds,liang2022metashift,koh2021wilds}, spurious correlation~\cite{srivastava2020robustness,agarwal2020towards}, etc.  Meanwhile, the corresponding methods in these fields may also inspire the future research in GLT, \eg, EIIL~\cite{creager2021environment}, GEORGE~\cite{sohoni2020no} and SDB~\cite{idrissi2022simple} can be used to construct better environments or baselines. Compared with these fields, the proposed GLT benchmarks provide a unified formulation and benchmarks for both class-wise and attribute-wise imbalances.



\section{Generalized Long-Tailed Classification}
\label{sec:3}

Previous LT methods~\cite{ren2020balanced,menon2020long} formulate the classification model as $p(Y|X)$, predicting the label $Y$ from the input image $X$, which can be further decomposed into $p(Y|X) \propto p(X|Y)\cdot p(Y)$~\cite{menon2020long,ren2020balanced}. This formulation identifies the cause of class-wise bias as $p(Y)$, so it can be elegantly solved by Logit Adjustment~\cite{menon2020long}. However, such a formulation is based on a strong assumption that the distribution of $p(X|Y)$ won't change in different domains, \ie, $p_{\text{train}}(X|Y) = p_{\text{test}}(X|Y)$, which cannot be guaranteed in real-world applications. Next, we will provide an in-depth analysis of why this over-simplified view fails to explain all biases.


\subsection{Problem Formulation} 


Recent studies~\cite{mirza2014conditional,besserve2018counterfactuals,wang2021self} demonstrate that an image of object $X$ can be fully described or generated by its class and a list of attributes. That is to say, each $X$ is generated by a set of underlying $(z_c, z_a)$\footnote{In this paper, $z_c$ and $z_a$ stand for \textbf{all} class-specific components and variant attributes, respectively, but we use a single variable to represent them in the following examples, \eg, $z_c=feather$ and $z_a=brown$, for simplicity.}, where the class-specific components $z_c$ are the invariant factors that enable the existence of the robust classification and the attribute-related variables $z_a$ are domain-specific knowledge that have inconsistent distributions. This formulation only assumes the invariance of a subset features $z_c$ rather than the entire $X$, \ie, $p_{\text{train}}(z_c|Y) = p_{\text{test}}(z_c|Y)$. Therefore, we can follow the Bayes theorem~\cite{stone2013bayes} to convert the classification model $p(Y|X)=p(Y|z_c,z_a)$ into the following formula: 
\begin{equation}
    \label{eq.1}
    p(Y=k|z_c, z_a) = \frac{p(z_c|Y=k)}{p(z_c)} \cdot \underbrace{\frac{p(z_a|Y=k,z_c)}{p(z_a|z_c)}}_{attribute\;bias} \cdot \underbrace{p(Y=k)}_{class\;bias},
\end{equation}
where invariant components $z_c$ only depend on $Y$; descriptive attributes $z_a$ that vary across instances may depend on both $Y$ and $z_c$. We generally consider $p(z_c, z_a) = p(z_a|z_c)\cdot p(z_c)$ WITHOUT introducing any independence assumption. Note that we also DO NOT impose the disentanglement assumption that a perfect feature vector $\mathbf{z}=[z_c;z_a]$ with separated $z_c$ and $z_a$ can be obtained, as the disentanglement is a challenging task on its own~\cite{locatello2019challenging}. Otherwise, we only need to conduct a simple feature selection to obtain the ideal classification model.

The reason why we need Eq.~\eqref{eq.1} to replace the simple $p(Y|X)$ is because unlike those tasks that indeed require both $z_c$ and $z_a$, \eg, the image captioning~\cite{you2016image} or the segmentation task~\cite{he2017mask}, the classification task merely relies on the class-specific components $z_c$ of an image $X$, regardless of its varying attributes $z_a$.  Therefore, the former formulation $p(Y|X)$ over-simplifies the problem by ignoring the different roles between $z_c$ and $z_a$ during classification.

\begin{wrapfigure}{r}{70mm}
   \includegraphics[width=70mm]{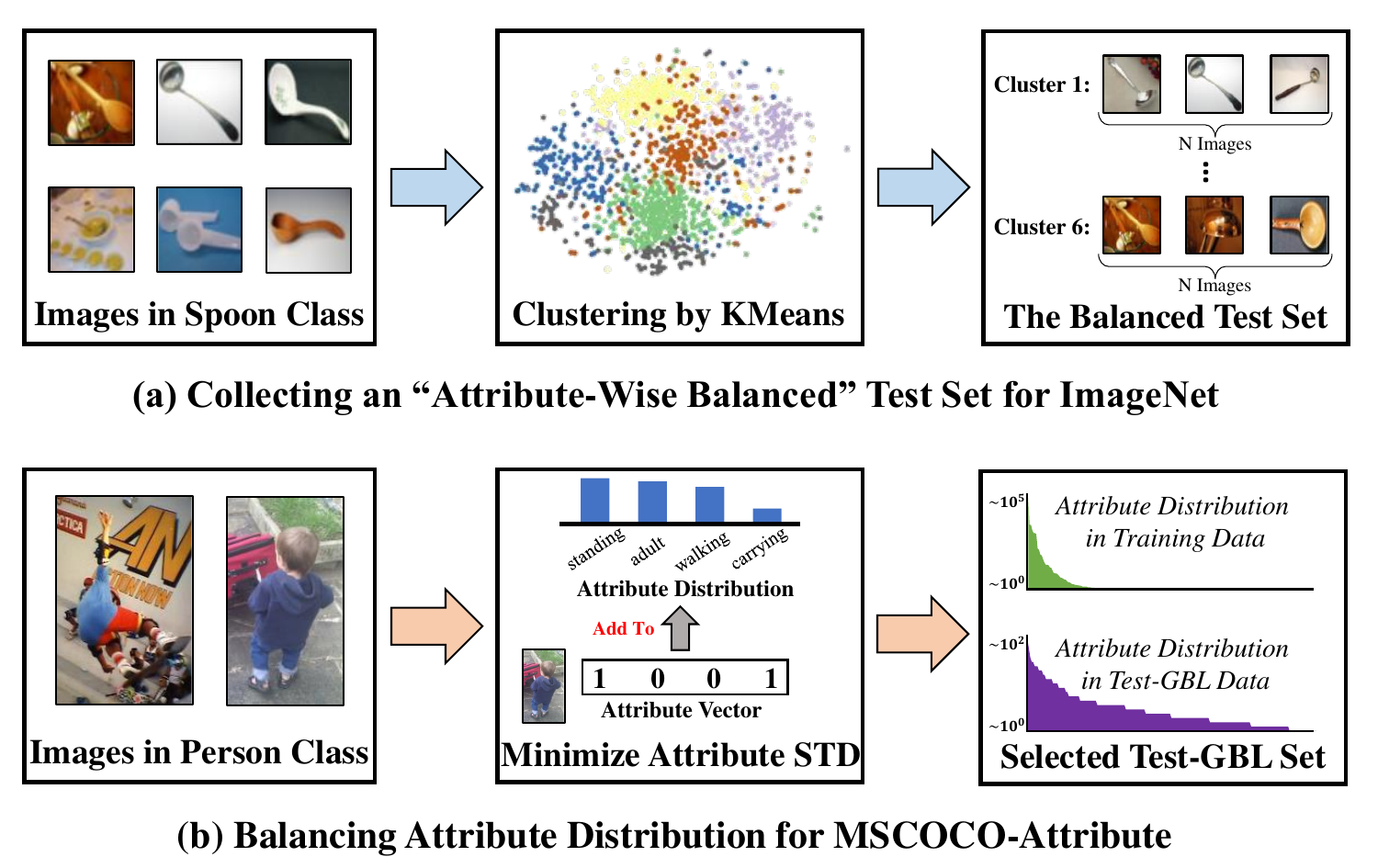}
   \caption{Examples of how to balance the attribute distribution for the Test-GBL evaluation environment in the proposed two benchmarks}
   \label{fig:4} 
\end{wrapfigure}


\noindent\textbf{Class Bias:} in class-wise LT~\cite{zhang2021deep},  the distribution of $p(Y)$ is considered as the main cause of the performance degradation. As $p(Y)$ can be explicitly calculated from the training data, the majority of previous LT methods directly alleviate its effect by class-wise adjustment~\cite{ren2020balanced,menon2020long} or re-balancing~\cite{kang2019decoupling,zhou2019bbn}. However, they fail to answer 
\textbf{I)} why the performance is also long-tailed within each class, and 
\textbf{II)} why tail images tend to be mis-classified as certain head classes with similar attributes.

\noindent\textbf{Attribute Bias:} the above Eq.\eqref{eq.1} extends the previous LT formulation by introducing the attribute bias caused by long-tailed $z_a$, which not only explains the cause of inconsistent performances within each class but also demonstrates how spurious correlations hurt the prediction. Intuitively, 
\textbf{I)} for the intra-class variation, if an attribute ``white''/``brown'' of the class-specific component ``feather'' is more frequent/rarer in ``dove'' than in other classes, \eg, $\frac{p(z_a=brown|Y=dove, z_c=feather)}{p(z_a=brown|z_c=feather)} < \frac{p(z_a=white|Y=dove, z_c=feather)}{p(z_a=white|z_c=feather)}$, a dove with brown feather will have lower confidence than doves with white feather, following Eq.~\eqref{eq.1}, \ie, $p(Y \myeq dove|z_c \myeq feather,z_a \myeq brown)$ $< p(Y \myeq dove|z_c \myeq feather,z_a \myeq white)$. 
\textbf{II)} Similarly, for spurious correlations, if a ``field'' background is more frequent in the class ``tractor'' than other vehicles with the same class-specific component ``wheel'', it will create false sense of correlation between ``field'' and ``tractor'' class, \ie, $\frac{p(z_a = field|Y = tractor, z_c=wheel)}{p(z_a = field|z_c=wheel)} >> 1$, resulting images of other vehicles with ``wheel'' blindly have larger $p(Y \myeq tractor|z_c \myeq wheel,z_a \myeq field)$ in the ``field'' background, \eg, predicting a ``harvester in field'' as the ``tractor''.

\noindent\textbf{Generalized Long-Tailed Distribution: } the proposed GLT asserts that both the conventional class distribution and the ever-overlooked attribute distribution are long-tailed in real-world dataset at scale. However, most of the previous LT benchmarks, ImageNet-LT~\cite{liu2019large}, Long-Tailed CIFAR-10/-100~\cite{zhou2019bbn}, or iNaturalist~\cite{van2018inaturalist}, are only capable of evaluating the class bias, underestimating the role of the attribute bias in the long tail challenge. To better study unbiased models that address both of biases at the same time, we introduce the following two GLT benchmarks and three evaluation protocols.



\subsection{GLT Benchmarks and Evaluation Protocols} 

In this section, we design two benchmarks, ImageNet-GLT and MSCOCO-GLT, for the proposed GLT challenge, where ImageNet-GLT is a long-tailed version of ImageNet~\cite{russakovsky2015imagenet} and MSCOCO-GLT is constructed from MSCOCO-Attribute~\cite{patterson2016coco}. Although there are explicit attribute annotations in MSCOCO-Attribute, we forbid the access of them during training to make the algorithm more general. After all, attributes are not exhaustively annotated. Meanwhile, ImageNet, like most of the other datasets, doesn't have any attribute annotation, so we cluster image features within each class into multiple ``pretext attributes'' using a pre-trained model~\cite{NEURIPS2019_9015}. These clusters can thus serve as annotations for implicit attributes. To systematically diagnose two kinds of biases in Eq.~\eqref{eq.1}, each benchmark is further organized into three evaluation protocols as follows:

\noindent\textbf{Class-wise Long Tail (CLT) Protocol:} same as the conventional LT, we first adopt a class-wise and attribute-wise LT training set, called \textbf{Train-GLT}, which can be easily sampled from ImageNet~\cite{russakovsky2015imagenet} and MSCOCO-Attribute~\cite{patterson2016coco} using a class-wise LT distribution. We don't need to intentionally ensure the attribute-wise imbalance as it's ubiquitous and inevitable in any real-world dataset, \eg, the distribution of MSCOCO-Attribute~\cite{patterson2016coco} in Fig.~\ref{fig:1}~(b). The corresponding \textbf{Test-CBL}, which is i.i.d. sampled within each class, is a class-wise balanced and attribute-wise long-tailed testing set. \textbf{(Train-GLT, Test-CBL)} with the same attribute distributions and different class distributions can thus evaluate the robustness against the class-wise long tail.

\noindent\textbf{Attribute-wise Long Tail (ALT) Protocol:} the training set \textbf{Train-CBL} of this protocol has the same number of images for each class and keeps the original long-tailed attribute distribution by i.i.d. sampling images within each class, so its bias only comes from the attribute. Meanwhile, \textbf{Test-GBL}, as the most important evaluation environment for GLT task, has to balance both class and attribute distributions. As illustrated in Fig.~\ref{fig:4}, Test-GBL for ImageNet-GLT samples equal number of images from each ``pretext attribute'' (\ie, feature clusters) and each class. Test-GBL for MSCOCO-GLT is a little bit tricky, because each object has multiple attributes, making strictly balancing the attribute distribution prohibitive. Hence, we select a fixed size of subset within each class that has the minimized standard deviation of attributes as the Test-GBL. As long as Test-GBL is relatively more balanced in attributes than Train-CBL, it can serve as a valid testing set for ALT protocol. In summary, \textbf{(Train-CBL, Test-GBL)} have the same class distributions and different attribute distributions.

\noindent\textbf{Generalized Long Tail (GLT) Protocol:} this protocol combines \textbf{(Train-GLT, Test-GBL)} from the above, so both class and attribute distributions are changed from training to testing. As the generalized evaluation protocol for the long-tailed challenge, an algorithm can only obtain satisfactory results when both class bias and attribute bias are well addressed by the final model.

\section{Invariant Feature Learning}
\label{sec:4}

\begin{wrapfigure}{R}{0.6\textwidth}
   \begin{minipage}[b]{1.0\linewidth}
   \centerline{\includegraphics[width=75mm]{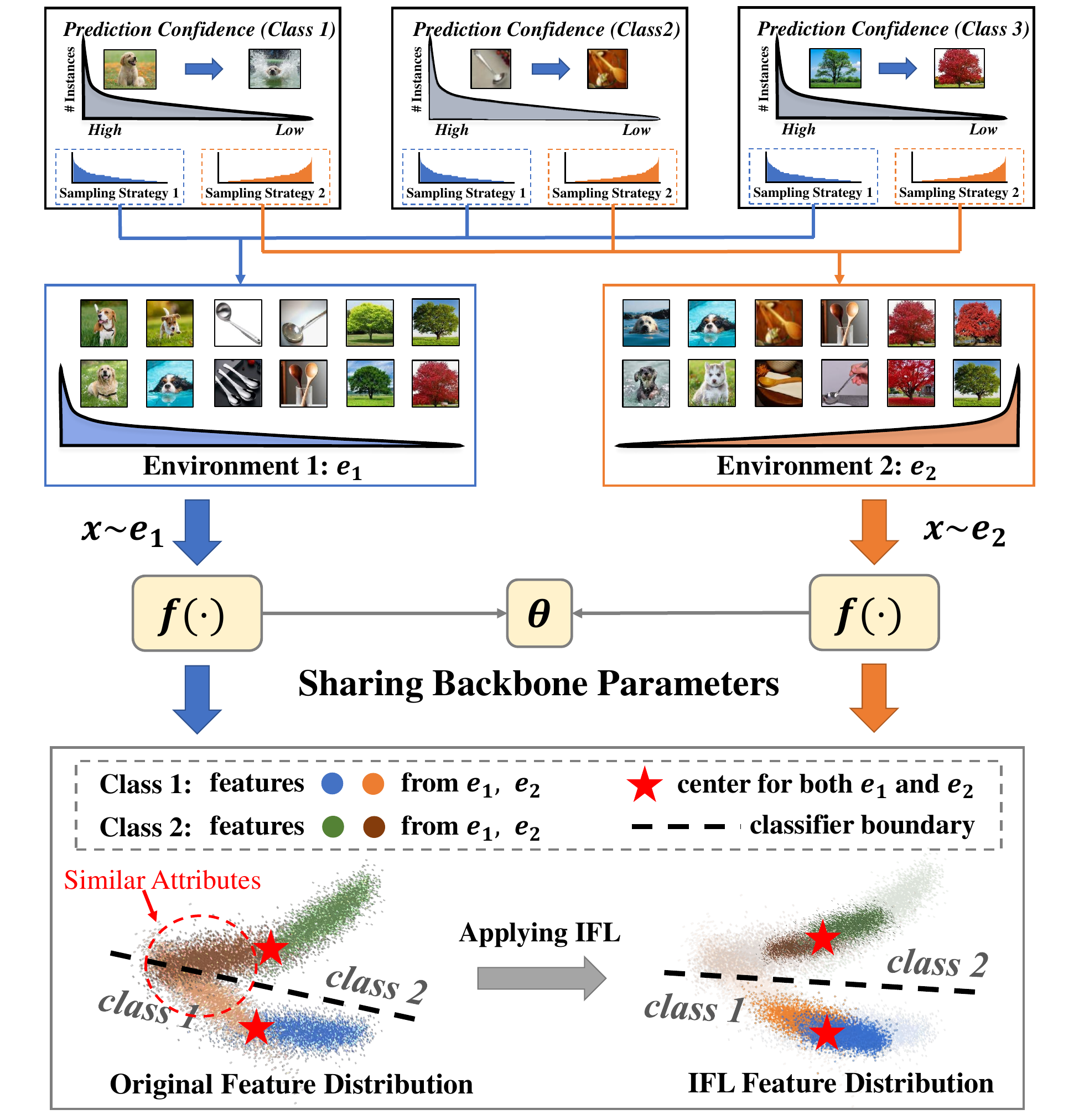}}
   \end{minipage}
   \caption{The proposed IFL that first applies different sampling strategies according to the prediction confidence within each class, then combines them to construct environments with diverse attribute distributions, and finally learns features invariant to the environment change by the IFL metric loss}
   \label{fig:5} 
\end{wrapfigure}

Prevalent LT studies~\cite{menon2020long,kang2019decoupling} mainly focus on designing balanced classifiers, as the $p(Y=k)$ in Eq.~\eqref{eq.1} is independent of input features. However, as we discussed in Section~\ref{sec:3},  our new attribute bias is caused by extracting the undesired $z_a$, so the overlooked feature learning is the key. Therefore, the proposed method aims to learn an improved feature backbone that is complementary for previous balanced classifiers~\cite{kang2019decoupling,tang2020long,ren2020balanced,menon2020long}.

As we discussed in Section~\ref{sec:3}, the attribute bias term in Eq.~\eqref{eq.1} raises two problems: 
\textbf{I)} the inconsistent performance of $p(Y|z_c,z_a)$ within each class, and 
\textbf{II)} the spurious correlations between a non-robust attribute and a class.
To tackle these problems, we propose Invariant Feature Learning (IFL) that extends the center loss~\cite{wen2016centerloss} from the original Empirical Risk Minimization (ERM) to its Invariant Risk Minimization (IRM)~\cite{arjovsky2019invariant} version, which forces the backbone to focus on learning features that are invariant across a set of environments.

%

The rationale behind the IFL is that $\frac{p(z_c|Y \myeq k)}{p(z_c)}$ is consistent in all environments while $\frac{p(z_a|Y=k,z_c)}{p(z_a|z_c)}$ is not. If images of $Y \myeq k$ have the same feature center in all environments, it means that the extracted features are more related to invariant $z_c$ instead of $z_a$. The overall IFL framework is summarized in Fig.~\ref{fig:5}. 



\subsection{Environment Construction}


After warming-up the model with the vanilla cross-entropy loss for several epochs, we obtain an initial model with imperfect predictions. Since $\frac{p(z_c|Y \myeq k)}{p(z_c)}$ and $p(Y \myeq k)$ are constants within a given class $Y \myeq k$, the variation of $p(Y \myeq k|z_c,z_a)$ is directly proportional to $\frac{p(z_a|Y=k,z_c)}{p(z_a|z_c)}$. It allows us to use the prediction confidence on the ground-truth class $p(Y \myeq k|X$ in $k)$ as the indicator to sample diverse attribute distributions in the training set (See Appendix for more discussions). As illustrated in Fig.~\ref{fig:5}, samples collected from each class using the same sampling strategy will then be combined together to form different environments in $\mathcal{E}$. Each sampling strategy has a unique distribution of $\frac{p(z_a|Y=k,z_c)}{p(z_a|z_c)}$. The environments are periodically updated during the training phase.

In general, two environments are considered to be sufficient to learn the robustness against the environment change~\cite{arjovsky2019invariant}, which is also supported by our experiments on Table~\ref{tab:4}. Specifically, one environment directly samples each instance uniformly inside each class, \ie, the na\"ive i.i.d. sampling. The other environment is sampled based on the $(1-p(Y=k|z_c,z_a))^\beta$, where $\beta$ automatically adjusts the new attribute distribution by up-sampling $20\%$ images from class $Y \myeq k$ with lowest $p(Y=k|z_c,z_a)$ to reach $80\%$ population of the class $Y \myeq k$ in the new environment, following the Pareto Principle~\cite{reed2001pareto}. 

The pseudo code of the algorithm is provided in the Appendix.


\subsection{IFL Metric Loss}

After obtaining a set of diverse training environments $\mathcal{E}$, the goal of the proposed IFL can thus be formulated as the following optimization problem:
\begin{equation}
    \label{eq.2}
    \begin{split}
         \min_{\theta, w} \; & \sum_{e\in\mathcal{E}} \sum_{i\in e} L_{cls}(f(x_i^e;\theta), y_i^e; w),\\
        \text{subject to} \; & \theta \in \arg\min_\theta \; \sum_{e\in\mathcal{E}} \sum_{i \in e} ||  f(x_i^e;\theta) - C_{y_i^e} ||_2, 
    \end{split}
\end{equation}
where $\theta$ and $w$ are learnable parameters for the backbone and classifier, respectively; $(x_i^e, y_i^e)$ are $i$-th $($image, label$)$ pair in the training environment $e\in \mathcal{E}$; $f(x_i^e;\theta)$ is the backbone extracting feature from $x_i^e$; $L_{cls}(f(x_i^e;\theta), y_i^e; w)$ is the cross-entropy loss;  $C_{y_i^e}$ is the mean feature of corresponding class $y_i^e$ across all environments in $\mathcal{E}$.\footnote{We follow the center loss~\cite{wen2016centerloss} to implement $C_{y_i^e}$ as the moving average for efficiency.}


The above Eq.~\eqref{eq.2} aims to optimize the model by the classification loss $L_{cls}$ under the constraint that the intra-class variation of features across all environments is also minimized, suppressing the learning of $z_a$ that causes non-robust $\frac{p(z_a|Y=k,z_c)}{p(z_a|z_c)}$. So the overall training objective is thus defined as $L= L_{cls} + \alpha \cdot L_{IFL}$, where $L_{IFL} = ||  f(x_i^e;\theta) - C_{y_i^e} ||_2$ is a metric loss that ensures the above constraint and $\alpha$ is the trade-off parameter. 

As the generalized IRM version of the center loss~\cite{wen2016centerloss}, the proposed IFL significantly boosts the GLT performance from its ERM counterpart using single environment. It's because class centers are also biased under the long-tailed attributes, so the original center loss would inevitably under-represent images with rare attributes.  More experimental analyses are given in Section~\ref{sec:5.5}.


\section{Experiments}
\label{sec:5}

\subsection{Datasets and Metrics} 


\noindent\textbf{ImageNet-GLT} is a long-tailed version of the ImageNet~\cite{russakovsky2015imagenet}, where CLT and GLT protocols share the same training set Train-GLT with 113k samples over 1k classes. ALT protocol adopts a class-wise balanced Train-CBL with 114k images. The evaluation splits \{Val, Test-CBL, Test-GBL\} have \{30k, 60k, 60k\} samples, respectively. The number of images for each class in Train-GLT ranges from 570 to 4, while all classes have 114 samples in Train-CBL. To collect the attribute-wise balanced Test-GBL, images from each class were clustered into 6 groups by KMeans using a pre-trained ResNet50 model~\cite{NEURIPS2019_9015} and we sampled 10 images per group and class. Following \cite{liu2019large,kang2019decoupling}, all testing sets are also split into 3 subsets by the class frequency: Many$_C$ with \#sample $>$ 100, Medium$_C$ with 100 $\geq$ \#sample $\geq$ 20, and Few$_C$ with \#sample $<$ 20. We further split them into 3 subsets by attribute groups: Many$_A$, Medium$_A$, and Few$_A$ with images from the most/medium/least frequent 2 clusters of all classes, respectively.

\begin{table*}[t!]
\centering
\caption{\textbf{Evaluation of CLT and GLT Protocols on ImageNet-GLT}: Accuracy (\textit{left in each cell}) and Precision (\textit{right in each cell}) are reported. All methods are re-implemented under the same codebase with ResNext-50 backbone}
\scalebox{0.55}
{
\begin{tabu}{c| c |[1.5pt]c |c |c |c |[1.5pt]c |c |c |c }
\hline
\hline
\multicolumn{2}{c|[1.5pt]}{Methods} & \multicolumn{4}{c|[1.5pt]}{\textbf{Class-Wise Long Tail (CLT) Protocol}} & \multicolumn{4}{c}{\textbf{Generalized Long Tail (GLT) Protocol}} \\ 
\hline 
\multicolumn{2}{c|[1.5pt]}{\textbf{$<$ Accuracy $\vert$ Precision $>$}} & Many$_C$ & Medium$_C$ & Few$_C$ & Overall & Many$_C$ & Medium$_C$ & Few$_C$ & Overall\\ 
\hline 

\multirow{13}{*}{{\rotatebox{90}{\textbf{Re-balance}}}} 

& Baseline & 59.34 $\vert$ 39.08 & 36.95 $\vert$ 52.87 & 14.39 $\vert$ 56.65 & 42.52 $\vert$ 47.92 & 50.98 $\vert$ 32.90 & 28.49 $\vert$ 44.72 & 10.28 $\vert$ 49.11 & 34.75 $\vert$ 40.65 \\
& cRT~\cite{kang2019decoupling} & 56.55 $\vert$ 45.79 & 42.89 $\vert$ 46.23 & 26.67 $\vert$ 41.47 & 45.92 $\vert$ 45.34 & 48.02 $\vert$ 38.40 & 34.16 $\vert$ 38.07 & 19.92 $\vert$ 33.50 & 37.57 $\vert$ 37.51 \\
& LWS~\cite{kang2019decoupling} & 55.38 $\vert$ 46.67 & 43.91 $\vert$ 46.87 & 30.11 $\vert$ 40.92 & 46.43 $\vert$ 45.90 & 47.15 $\vert$ 39.16 & 34.88 $\vert$ 38.68 & 22.56 $\vert$ 32.88 & 37.94 $\vert$ 38.01 \\
& Deconfound-TDE~\cite{tang2020long} & 54.94 $\vert$ 49.27 & 43.18 $\vert$ 43.91 & 28.64 $\vert$ 33.40 & 45.70 $\vert$ 44.48 & 46.87 $\vert$ 42.39 & 34.43 $\vert$ 35.77 & 22.11 $\vert$ 26.30 & 37.56 $\vert$ 37.00 \\
& BLSoftmax~\cite{ren2020balanced} & 55.60 $\vert$ 48.19 & 42.74 $\vert$ 47.27 & 28.79 $\vert$ 38.14 & 45.79 $\vert$ 46.27 & 47.15 $\vert$ 40.89 & 33.48 $\vert$ 39.11 & 21.10 $\vert$ 27.50 & 37.09 $\vert$ 38.08 \\
& Logit-Adj~\cite{menon2020long} & 54.55 $\vert$ 49.70 & 44.40 $\vert$ 45.05 & 31.53 $\vert$ 36.04 & 46.53 $\vert$ 45.56 & 45.94 $\vert$ 41.97 & 35.15 $\vert$ 36.63 & 24.07 $\vert$ 28.59 & 37.80 $\vert$ 37.56 \\
& BBN~\cite{zhou2019bbn} & 61.64 $\vert$ 42.74 & 43.80 $\vert$ 54.44 & 13.94 $\vert$ 55.12 & 46.46 $\vert$ 49.86 & 52.41 $\vert$ 35.58 & 34.31 $\vert$ 46.38 & 10.06 $\vert$ 44.43 & 37.91 $\vert$ 41.77 \\

& LDAM~\cite{cao2019ldam} & 59.05 $\vert$ 45.39 & 43.23 $\vert$ 48.80 & 24.44 $\vert$ 44.99 & 46.74 $\vert$ 46.86 & 51.02 $\vert$ 38.78 & 34.13 $\vert$ 40.39 & 18.46 $\vert$ 35.91 & 38.54 $\vert$ 39.08 \\

& (ours) Baseline + IFL & \textbf{62.71} $\vert$ 42.98 & 40.10 $\vert$ \textbf{56.83} & 18.92 $\vert$ \textbf{61.92} & 45.97 $\vert$ \textbf{52.06} & \textbf{54.09} $\vert$ 36.74 & 31.73 $\vert$ \textbf{49.03} & 13.62 $\vert$ \textbf{51.42} & 37.96 $\vert$ \textbf{44.47} \\
& (ours) cRT + IFL & 61.27 $\vert$ 45.84 & 43.96 $\vert$ 51.67 & 24.32 $\vert$ 53.64 & 47.94 $\vert$ 49.63 & 52.75 $\vert$ 39.11 & 35.14 $\vert$ 43.36 & 17.92 $\vert$ 43.35 & 39.60 $\vert$ 41.65 \\
& (ours) LWS + IFL & 61.50 $\vert$ 45.43 & 43.79 $\vert$ 52.85 & 23.86 $\vert$ 55.58 & 47.89 $\vert$ 50.29 & 53.21 $\vert$ 38.92 & 34.99 $\vert$ 44.44 & 17.42 $\vert$ 45.90 & 39.64 $\vert$ 42.45 \\
& (ours) BLSoftmax + IFL & 58.00 $\vert$ 53.70 & 44.70 $\vert$ 51.73 & 33.49 $\vert$ 37.58 & 48.34 $\vert$ 50.39 & 49.92 $\vert$ 46.86 & 36.11 $\vert$ 44.31 & 25.71 $\vert$ 32.01 & 40.08 $\vert$ 43.48 \\
& (ours) Logit-Adj + IFL & 56.96 $\vert$ \textbf{56.22} & \textbf{46.54} $\vert$ 50.10 & \textbf{36.88} $\vert$ 33.29 & \textbf{49.26} $\vert$ 50.02 & 48.25 $\vert$ \textbf{49.17} & \textbf{37.50} $\vert$ 41.65 & \textbf{29.00} $\vert$ 25.77 & \textbf{40.52} $\vert$ 42.28 \\

\tabucline[1.5pt]{-}

\multirow{4}{*}{{\rotatebox{90}{\small{\textbf{Augment}}}}} 
& Mixup~\cite{zhang2018mixup} & 59.68 $\vert$ 37.96 & 30.83 $\vert$ 55.74 & 7.09 $\vert$ 34.33 & 38.81 $\vert$ 45.41 & 51.04 $\vert$ 31.85 & 23.10 $\vert$ 47.25 & 4.94 $\vert$ 22.88 & 31.55 $\vert$ 37.44 \\
& RandAug~\cite{cubuk2020randaugment} & 64.96 $\vert$ 42.63 & 40.30 $\vert$ 59.10 & 15.20 $\vert$ 56.60 & 46.40 $\vert$ 52.13 & 56.36 $\vert$ 35.97 & 31.43 $\vert$ 51.13 & 10.36 $\vert$ 48.92 & 38.24 $\vert$ 44.74 \\
& (ours) Mixup + IFL & 67.71 $\vert$ 47.77 & 45.87 $\vert$ 62.58 & 24.71 $\vert$  \textbf{67.77} & 51.43 $\vert$ 57.44 & 59.36 $\vert$ 40.95 & 36.77 $\vert$ 54.67 & 18.06 $\vert$ 55.10 & 43.00 $\vert$ 49.25 \\
& (ours) RandAug + IFL & \textbf{69.35} $\vert$ \textbf{49.42} & \textbf{48.05} $\vert$ \textbf{63.19} & \textbf{26.92} $\vert$ 66.04 & \textbf{53.40} $\vert$ \textbf{58.11} & \textbf{60.79} $\vert$ \textbf{42.41} & \textbf{39.07} $\vert$ \textbf{55.15} & \textbf{20.04} $\vert$ \textbf{57.90} & \textbf{44.90} $\vert$ \textbf{50.47} \\

\tabucline[1.5pt]{-}

\multirow{4}{*}{{\rotatebox{90}{\small{\textbf{Ensemble}}}}} 
& TADE~\cite{zhang2021test} & 58.44 $\vert$ \textbf{56.38} & 48.01 $\vert$ 51.41 & \textbf{36.60} $\vert$ 41.08 & 50.47 $\vert$ 51.85 & 50.29 $\vert$ \textbf{49.25} & 38.74 $\vert$ 43.74 & \textbf{27.99} $\vert$ 31.75 & 41.75 $\vert$ 44.15 \\
& RIDE~\cite{wang2020long} & 64.04 $\vert$ 51.91 & 48.66 $\vert$ 53.21 & 30.44 $\vert$ 46.25 & 52.08 $\vert$ 51.65 & 55.47 $\vert$ 44.55 & 38.65 $\vert$ 44.26 & 22.80 $\vert$ 37.26 & 43.00 $\vert$ 43.32 \\
& (ours) TADE + IFL & 61.71 $\vert$ 55.59 & 48.87 $\vert$ 53.42 & 34.02 $\vert$ 40.93 & 51.78 $\vert$ 52.41 & 53.75 $\vert$ 48.73 & 39.90 $\vert$ 45.28 & 26.77 $\vert$ 35.34 & 43.47 $\vert$ 45.17 \\
& (ours) RIDE + IFL & \textbf{65.68} $\vert$ 54.13 & \textbf{50.82} $\vert$ \textbf{56.22} & 31.91 $\vert$ \textbf{52.10} & \textbf{53.93} $\vert$ \textbf{54.76} & \textbf{57.84} $\vert$ 47.00 & \textbf{41.80} $\vert$ \textbf{48.65} & 24.63 $\vert$ \textbf{42.96} & \textbf{45.64} $\vert$ \textbf{47.14} \\

\hline
\hline
\end{tabu}
}
\label{tab:1}
\end{table*}

\noindent\textbf{MSCOCO-GLT} is a long-tailed subset of MSCOCO-Attribute~\cite{patterson2016coco,lin2014microsoft} with 196 different attributes. We cropped each object with multi-label attributes as independent images. Under CLT and GLT protocols, we have \{Train-GLT, Val, Test-CBL, Test-GBL\} with \{144k, 2.9k, 5.8k, 5.8k\} images over 29 classes, where the number of samples for each class ranges from 61k to 0.3k. The ALT protocol has \{32k, 1.4k, 2.9k\} images for \{Train-CBL, Val, Test-GBL\}. Since attributes usually co-occur with each other in one object, we cannot construct Many$_A$, Medium$_A$, and Few$_A$ subsets the same as ImageNet-GLT, so we directly report the overall performance for MSCOCO-GLT. Note that attribute annotations are only used to construct Test-GBL, and they are not released in the training data.


\noindent\textbf{Evaluation Metrics.} The top-1 accuracy is commonly adopted as the only metric in the conventional LT studies, yet, it cannot reveal the limitation of precision-accuracy trade-off. Therefore, in GLT classification, we report both Accuracy: $ \frac{\#Correct Predictions}{\#All Samples}$, which is equal to Top-1 Recall in the class-wise balanced test sets~\cite{zhu2021cross}, and Precision: $\frac{1}{\#class} \cdot \sum_{class} \frac{\#Correct Predictions}{\#Samples Predicted As This Class}$ to better evaluate the effectiveness of algorithms.

\subsection{Investigated LT Algorithms} 

As a general feature learning method to deal with the attribute bias, the proposed IFL can be integrated into most prevalent LT methods. We followed Zhang~\etal~\cite{zhang2021deep} to summarize the investigated LT algorithms into three categories: 1) one-/two-stage re-balancing, 2) augmentation, and 3) module improvement.

\begin{wraptable}{r}{70mm}
\centering
\caption{\textbf{Evaluation of ALT Protocol on ImageNet-GLT}}
\scalebox{0.5}
{
\begin{tabu}{c| c |[1.5pt]c |c |c |c}
\hline
\hline
\multicolumn{2}{c|[1.5pt]}{Methods} & \multicolumn{4}{c}{Attribute-Wise Long Tail (ALT) Protocol} \\ 
\hline 
\multicolumn{2}{c|[1.5pt]}{\textbf{$<$ Accuracy $\vert$ Precision $>$}} & Many$_A$ & Medium$_A$ & Few$_A$ & Overall \\ 
\hline 

\multirow{12}{*}{{\rotatebox{90}{\small{\textbf{Re-balance}}}}} 

&Baseline & 56.95 $\vert$ 55.83 & 40.11 $\vert$ 39.17 & 28.12 $\vert$ 28.16 & 41.73 $\vert$ 41.74 \\

&cRT~\cite{kang2019decoupling} & 57.45 $\vert$ 56.28 & 39.72 $\vert$ 38.65 & 27.58 $\vert$ 27.35 & 41.59 $\vert$ 41.43 \\

&LWS~\cite{kang2019decoupling} & 56.95 $\vert$ 55.85 & 40.11 $\vert$ 39.30 & 28.03 $\vert$ 27.98 & 41.70 $\vert$ 41.71 \\

&Deconfound-TDE~\cite{tang2020long} & 57.10 $\vert$ 56.58 & 39.80 $\vert$ 40.08 & 27.29 $\vert$ 27.96 & 41.40 $\vert$ 42.36 \\

&BLSoftmax~\cite{ren2020balanced} & 56.48 $\vert$ 55.56 & 39.81 $\vert$ 38.96 & 27.64 $\vert$ 27.60 & 41.32 $\vert$ 41.37 \\

&BBN~\cite{zhou2019bbn} & 60.90 $\vert$ 60.17 & 41.08 $\vert$ 40.81 & 27.79 $\vert$ 28.26 & 43.26 $\vert$ 43.86 \\

&LDAM~\cite{cao2019ldam} & 59.04 $\vert$ 56.51 & 40.96 $\vert$ 39.21 & 27.96 $\vert$ 27.22 & 42.66 $\vert$ 41.80 \\

&(ours) Baseline + IFL & \textbf{61.38} $\vert$ \textbf{60.78} & \textbf{44.79} $\vert$ \textbf{44.21} & \textbf{31.49} $\vert$ \textbf{31.98} & \textbf{45.89} $\vert$ \textbf{46.42} \\

&(ours) cRT + IFL & 61.12 $\vert$ 60.25 & 44.26 $\vert$ 43.65 & 31.02 $\vert$ 31.31 & 45.47 $\vert$ 45.81 \\

&(ours) LWS + IFL & 61.19 $\vert$ 60.45 & 44.66 $\vert$ 44.07 & 31.43 $\vert$ 31.91 & 45.76 $\vert$ 46.25 \\

&(ours) BLSoftmax + IFL & 60.19 $\vert$ 59.46 & 43.54 $\vert$ 43.14 & 30.85 $\vert$ 31.46 & 44.86 $\vert$ 45.43 \\

\tabucline[1.5pt]{-}

\multirow{4}{*}{{\rotatebox{90}{\small{\textbf{Augment}}}}} 

&Mixup~\cite{zhang2018mixup} & 58.71 $\vert$ 58.04 & 40.09 $\vert$ 38.99 & 27.52 $\vert$ 27.54 & 42.11 $\vert$ 42.42 \\

&RandAug~\cite{cubuk2020randaugment} & 62.35 $\vert$ 61.25 & 45.04 $\vert$ 44.27 & 31.47 $\vert$ 31.26 & 46.29 $\vert$ 46.32 \\

&(ours) Mixup + IFL & 65.90 $\vert$ 65.88 & 49.43 $\vert$ 49.43 & 35.40 $\vert$ 35.89 & 50.24 $\vert$ 51.04 \\

&(ours) RandAug + IFL & \textbf{67.39} $\vert$ \textbf{66.81} & \textbf{51.55} $\vert$ \textbf{51.28} & \textbf{37.47} $\vert$ \textbf{37.97} & \textbf{52.14} $\vert$ \textbf{52.74} \\

\tabucline[1.5pt]{-}

\multirow{4}{*}{{\rotatebox{90}{\small{\textbf{Ensemble}}}}} 

&TADE~\cite{zhang2021test} & 62.63 $\vert$ 61.91 & 45.84 $\vert$ 45.21 & 32.82 $\vert$ 32.82 & 47.10 $\vert$ 47.32 \\

&RIDE~\cite{wang2020long} & 63.48 $\vert$ 61.42 & 45.62 $\vert$ 44.16 & 32.59 $\vert$ 32.26 & 47.24 $\vert$ 46.67 \\

&(ours) TADE + IFL & 63.50 $\vert$ 62.67 & 48.03 $\vert$ 47.32 & 34.69 $\vert$ 34.52 & 48.74 $\vert$ 48.78 \\

&(ours) RIDE + IFL & \textbf{67.54} $\vert$ \textbf{67.13} & \textbf{51.92} $\vert$ \textbf{51.72} & \textbf{37.84} $\vert$ \textbf{38.46} & \textbf{52.44} $\vert$ \textbf{53.17} \\

\hline
\hline
\end{tabu}
}
\label{tab:2}
\end{wraptable}

For re-balancing approaches, we studied two-stage re-sampling methods \textbf{cRT}~\cite{kang2019decoupling} and \textbf{LWS}~\cite{kang2019decoupling}, post-hoc distribution adjustment \textbf{Deconfound-TDE}~\cite{tang2020long} and Logit Adjustment (\textbf{Logit-Adj})~\cite{menon2020long}, multi-branch models with diverse sampling strategies like \textbf{BBN}~\cite{zhou2019bbn}, and re-weighting loss functions like Balanced Softmax (\textbf{BLSoftmax})~\cite{ren2020balanced} and \textbf{LDAM}~\cite{cao2019ldam}.

For augmentation approaches, we empirically noticed that some common data augmentation methods are more general and effective than other long-tailed transfer learning methods~\cite{liu2020deep}, so we adopted \textbf{Mixup}~\cite{zhang2018mixup} and Random Augmentation (\textbf{RandAug})~\cite{cubuk2020randaugment} in our experiments.

For module improvement, we followed the recent trend of ensemble learning~\cite{wang2020long} like \textbf{RIDE}~\cite{wang2020long} and \textbf{TADE}~\cite{zhang2021test}, which are proved to be state-of-the-art models in LT classification that are capable of improving both head and tail categories at the same time. Both of them used the trident version of ResNext-50~\cite{xie2017aggregated} developed by RIDE~\cite{wang2020long}, which is denoted as RIDE-50 in Table~\ref{tab:4}.

Implementation details of the proposed IFL and the baseline methods are given in Appendix.

\subsection{Comparisons with LT Line-up}

We evaluate CLT and GLT protocols for ImageNet-GLT in Table~\ref{tab:1}. The corresponding ALT protocol is reported in Table~\ref{tab:2}. All three protocols for MSCOCO-GLT are shown in Table~\ref{tab:3}.

\noindent\textbf{CLT Protocol (conventional long-tailed classification): } although the proposed IFL is mainly designed to tackle the long-tailed intra-class attributes, it also boost all prevalent LT methods in the conventional CLT protocol. It's because IFL also prevents tail images from being mis-classified as head classes by eliminating the spurious correlation made by attributes.

\begin{wraptable}{r}{70mm}
\centering
\caption{\textbf{Evaluation on MSCOCO-GLT:} overall performances are reported}
\scalebox{0.6}
{
\begin{tabu}{c| c |[1.5pt]c |c |[1.5pt]c}
\hline
\hline
\multicolumn{2}{c|[1.5pt]}{Protocols} & CLT & GLT & ALT \\ 
\hline
\multicolumn{2}{c|[1.5pt]}{\textbf{$<$ Accuracy $\vert$ Precision $>$}} & Overall & Overall & Overall \\ 
\hline 

\multirow{13}{*}{{\rotatebox{90}{\small{\textbf{Re-balance}}}}} 

&Baseline &  72.34 $\vert$ 76.61 & 63.79 $\vert$ 70.52 &  50.17 $\vert$ 50.94 \\

&cRT~\cite{kang2019decoupling} &  73.64 $\vert$ 75.84 & 64.69 $\vert$ 68.33 & 49.97 $\vert$ 50.37 \\

&LWS~\cite{kang2019decoupling} &  72.60 $\vert$ 75.66 & 63.60 $\vert$ 68.81 & 50.14 $\vert$ 50.61 \\

&Deconfound-TDE~\cite{tang2020long} & 73.79 $\vert$ 74.90 & 66.07 $\vert$ 68.20 & 50.76 $\vert$ 51.68 \\

&BLSoftmax~\cite{ren2020balanced} & 72.64 $\vert$ 75.25 & 64.07 $\vert$ 68.59 & 49.72 $\vert$ 50.65 \\

& Logit-Adj~\cite{menon2020long}  & 75.50 $\vert$ 76.88 & 66.17 $\vert$ 68.35 &  50.17 $\vert$ 50.94 \\

&BBN~\cite{zhou2019bbn} &  73.69 $\vert$ 77.35 & 64.48 $\vert$ 70.20 & 51.83 $\vert$ 51.77 \\

&LDAM~\cite{cao2019ldam} & 75.57 $\vert$ 77.70 & 67.26 $\vert$ 70.70 & \textbf{55.52} $\vert$ \textbf{56.21} \\

&(ours) Baseline + IFL & 74.31 $\vert$ 78.90 & 65.31 $\vert$ \textbf{72.24} & 52.86 $\vert$ 53.49 \\

&(ours) cRT + IFL & 76.21 $\vert$ 79.11 & 66.90 $\vert$ 71.34 & 52.07 $\vert$ 52.85 \\

&(ours) LWS + IFL & 75.98 $\vert$ \textbf{79.18} & 66.55 $\vert$ 71.49 & 52.07 $\vert$ 52.90 \\

&(ours) BLSoftmax + IFL &  73.72 $\vert$ 77.08 & 64.76 $\vert$ 70.00 & 52.97 $\vert$ 53.52\\

&(ours) Logit-Adj + IFL & \textbf{77.16} $\vert$ 79.09 & \textbf{67.53} $\vert$ 70.18 & 52.86 $\vert$ 53.49 \\

\tabucline[1.5pt]{-}

\multirow{4}{*}{{\rotatebox{90}{\small{\textbf{Augment}}}}} 

&Mixup~\cite{zhang2018mixup} & 74.22 $\vert$ 78.61 & 64.45 $\vert$ 71.13 & 48.90 $\vert$ 49.53 \\

&RandAug~\cite{cubuk2020randaugment} & 76.81 $\vert$ 79.88 & 67.71 $\vert$ 72.73 & 53.69 $\vert$ 54.71 \\

&(ours) Mixup + IFL & 77.55 $\vert$ \textbf{81.78} & \textbf{68.83} $\vert$ \textbf{74.84} & 53.79 $\vert$ 54.60 \\

&(ours) RandAug + IFL & \textbf{77.71} $\vert$ 81.10 & 68.16 $\vert$ 73.97 & \textbf{56.62} $\vert$ \textbf{57.12} \\

\tabucline[1.5pt]{-}

\multirow{4}{*}{{\rotatebox{90}{\small{\textbf{Ensemble}}}}} 

&TADE~\cite{zhang2021test} & 76.22 $\vert$ 78.84 & 66.98 $\vert$ 71.22 & 54.93 $\vert$ 55.48 \\

&RIDE~\cite{wang2020long} & 78.29 $\vert$ 80.33 & 68.59 $\vert$ 72.20 & 58.90 $\vert$ 59.43 \\

&(ours) TADE + IFL & 76.53 $\vert$ 79.15 & 67.38 $\vert$ 72.42 & 56.76 $\vert$ 57.43 \\

&(ours) RIDE + IFL & \textbf{78.86} $\vert$ \textbf{80.70} & \textbf{69.09} $\vert$ \textbf{72.57} & \textbf{58.93} $\vert$ \textbf{59.84} \\

\hline
\hline
\end{tabu}
}
\label{tab:3}
\end{wraptable}

\noindent\textbf{GLT Protocol:} not surprisingly, we observe a significant performance decline for all methods from CLT protocol to GLT, as tackling the additional attribute bias is much more challenging than class bias. Meanwhile, IFL can still successfully improve various baselines. The decline on the proposed GLT reveals that general long tail is indeed not just the pure class-wise imbalance. 

\noindent\textbf{ALT Protocol:} as we expected, the majority of previous LT algorithms using re-balancing strategies failed to improve the robustness against the attribute-wise bias in this protocol. Therefore, their improvements on GLT protocol only came from the class-wise invariance. However, we noticed that the augmentation and ensemble approaches can improve all three protocols, making them good baselines for GLT as well. The main reason is that both augmentation and model ensemble also aim to improve the representation learning.

\subsection{Ablation Studies and Further Analyses}
\label{sec:5.5}

We further conducted a group of ablation studies on ImageNet-GLT in Table~\ref{tab:4} to address some common concerns. We also provide some further analyses to shed lights on the proposed GLT challenge.

\noindent\textbf{Q1:} \textbf{how about the baselines from the attribute-wise side?} \textbf{A1:} we also reported some popular methods like Focal loss~\cite{lin2017focal} for hard-example mining and LFF (Learning-from-failure)~\cite{nam2020learning} for domain generalization as the baselines solving the attribute bias. However, they didn't perform well using their default settings, proving the difficulty of the proposed GLT in real-world datasets.


\noindent\textbf{Q2:} \textbf{does the improvement come from the center loss?} \textbf{A2:}
the original ERM version of center loss~\cite{wen2016centerloss} can be considered as a special case of the proposed IFL with only one single environment. According to Table~\ref{tab:4}, using the vanilla center loss, \ie, \#Env=1 with IFL, as the additional constraint actually hurt all three protocols. It's because the center in a biased environment is also biased, \eg, the center of ``banana'' may possess 90\% of ``yellow'' attribute, which only makes the model more relying on spurious correlations. 

\begin{wraptable}{r}{70mm}
\centering
\caption{Ablation Studies on ImageNet-GLT, where overall results are reported; BLS, Focal, and IFF are balanced softmax loss~\cite{ren2020balanced}, focal loss~\cite{lin2017focal}, and learning from failure~\cite{nam2020learning}, respectively}
\scalebox{0.5}
{
\begin{tabu}{c |c |c |c |c |[2pt]c |c |c}
\hline
\hline
\multicolumn{5}{c|[2pt]}{Ablation Settings}  & \multicolumn{3}{c}{Evaluation Protocols} \\ 
\hline 
\#Env & Loss & IFL & Augment & Backbone & CLT Protocol & GLT Protocol & ALT Protocol \\ 
\hline 
1 & CE & - & - & ResNext-50 & 42.52 $\vert$ 47.92 & 34.75 $\vert$ 40.65 & 41.73 $\vert$ 41.74 \\
1 & Focal & - & - & ResNext-50 & 39.93 $\vert$ 46.99 & 32.52 $\vert$ 39.12 & 39.58 $\vert$ 39.85 \\
1 & LFF & - & - & ResNext-50 & 41.07 $\vert$ 45.79 & 33.84 $\vert$ 38.46 & 40.14 $\vert$ 40.58 \\
1 & CE & \ding{52} & - & ResNext-50 & 39.74 $\vert$ 47.06 & 32.82 $\vert$ 40.86 & 39.99 $\vert$ 41.38 \\
2 & IRM & - & - & ResNext-50 & 43.70 $\vert$ 48.06 & 36.03 $\vert$ 40.61 & 44.47 $\vert$ 44.60 \\
2 & CE & \ding{52} & - & ResNext-50 & 45.97 $\vert$ 52.06 & 37.96 $\vert$ 44.47 & 45.89 $\vert$ 46.42 \\
3 & CE & \ding{52} & - & ResNext-50 & 46.06 $\vert$ 52.81 & 38.32 $\vert$ 45.55 & 45.95 $\vert$ 46.43 \\

2 & BLS & \ding{52} & - & ResNext-50 & 48.34 $\vert$ 50.39 & 40.08 $\vert$ 43.48 & 44.86 $\vert$ 45.43 \\
2 & CE & \ding{52} & Mixup & ResNext-50 & 51.43 $\vert$ 57.44 & 43.00 $\vert$ 49.25 & 50.24 $\vert$ 51.04 \\
2 & CE & \ding{52} & RandAug & ResNext-50 & \textbf{53.40} $\vert$ \textbf{58.11} & \textbf{44.90} $\vert$ \textbf{50.47} & \textbf{52.14} $\vert$ \textbf{52.74} \\
\hline

1 & CE & - & - & RIDE-50 & 46.14 $\vert$ 52.98  & 38.25 $\vert$ 45.80 &  46.32 $\vert$ 46.56 \\
2 & CE & \ding{52} & - & RIDE-50 & 49.20 $\vert$ 54.64 & 41.35 $\vert$ 47.67 & 48.62 $\vert$ 48.62 \\

2 & TADE & \ding{52} & - & RIDE-50 & 51.78 $\vert$ 52.41 & 43.47 $\vert$ 45.17 & 48.74 $\vert$ 48.78 \\
2 & LDAM & \ding{52} & - & RIDE-50 & 53.93 $\vert$ 54.76 & 45.64 $\vert$ 47.14 & 52.44 $\vert$ 53.17 \\
2 & LDAM & \ding{52} & Mixup & RIDE-50 & 56.48 $\vert$ 57.67 & 47.54 $\vert$ 49.86 & 53.25 $\vert$ 54.27 \\
2 & LDAM & \ding{52} & RandAug & RIDE-50 & \textbf{58.70} $\vert$ \textbf{59.61} & \textbf{49.80} $\vert$ \textbf{51.62} &  \textbf{55.65} $\vert$ \textbf{55.81} \\

\hline
\hline
\end{tabu}
}
\label{tab:4}
\end{wraptable}

\noindent\textbf{Q3:} \textbf{how many environments are required?} \textbf{A3:} although it's crucial to have more than 1 environment, IRM~\cite{arjovsky2019invariant} asserts that two environments are enough to capture the invariance. We also found that additional environments only brought marginal improvement. 

\noindent\textbf{Q4:} \textbf{why not directly apply IRM loss?} \textbf{A4:} theoretically, the original IRM loss~\cite{arjovsky2019invariant} and the proposed IFL are supposed to have similar results, as they both embody the same spirit of learning invariance across environments. However, we empirically noticed that the original IRM loss has the convergence issue in real-world dataset. 2 out of 5 random seeds result NaN loss at some point during training.

\noindent\textbf{Q5: why do all models perform worse under ALT than GLT on MSCOCO-GLT?} \textbf{A5: } note that the GLT protocol is always harder than ALT or CLT. This weird phenomenon in MSCOCO-GLT is caused by the severe class-wise imbalance in MSCOCO-Attributes~\cite{patterson2016coco}: a single class ``person'' possess over 40\% of the training set, so Train-CBL has much less training samples than Train-GLT. It further proves the importance of long-tailed classification in real-world applications, as \textit{large long-tailed datasets are better than small balanced counterparts}.

\begin{figure}
  \includegraphics[width=\linewidth]{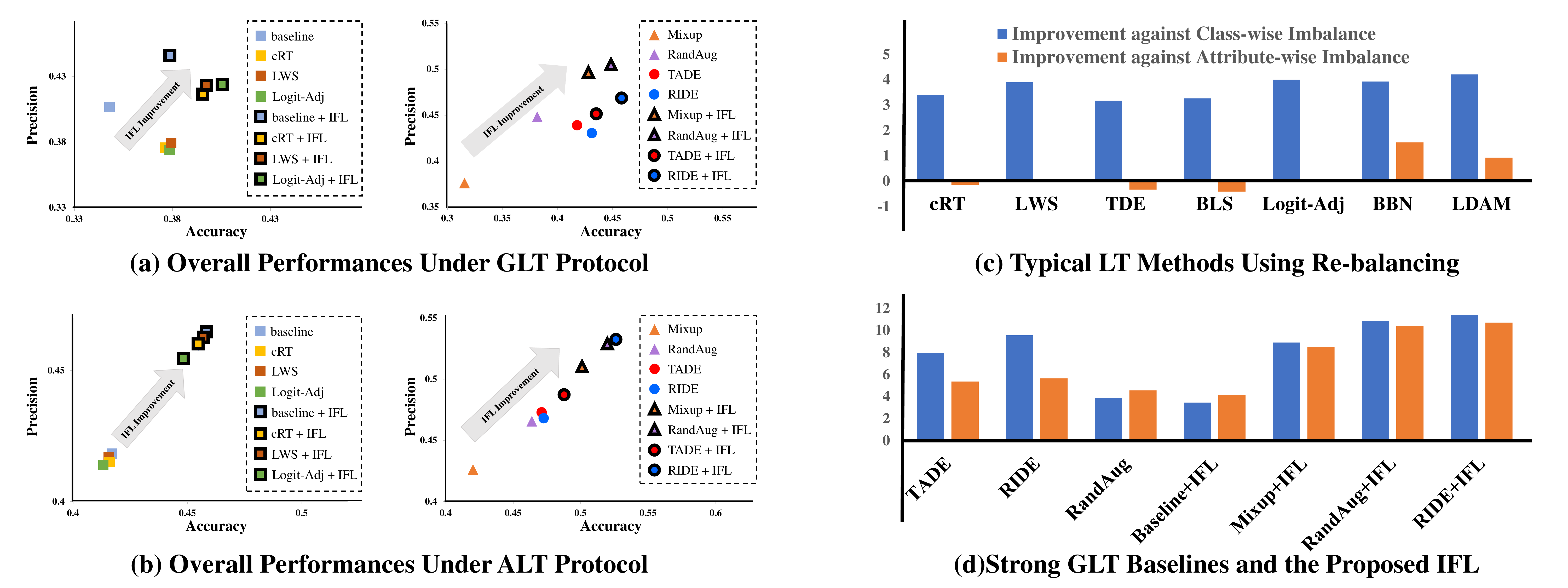}
  \caption{(a-b) The trending of precision and accuracy after applying the IFL; (c-d) GLT baselines will automatically improve class-wise LT, while conventional LT re-balancing algorithms won't improve the attribute-wise imbalance in GLT}
\label{fig:8}
\end{figure}

\noindent\textbf{Q6: what is the precision-accuracy trade-off problem~\cite{zhu2021cross}?} \textbf{A6:} as shown in Fig.~\ref{fig:8}~(a-b), class-wise re-balancing method like cRT, LWS, and Logit-Adj~\cite{kang2019decoupling,menon2020long} are playing with the precision-accuracy trade-off under GLT protocol and barely work in ALT protocol. It's because the attribute bias hurts both accuracy and precision by forming spurious correlations between classes. Tackling it should improve both of metrics at the same time.



\noindent\textbf{Q7: is there any other method that can be served as GLT baselines as well?} \textbf{A7:} we also found that the recent trend of improving both head and tail categories~\cite{wang2020long,zhang2021test,zhu2021cross}, though lack a formal definition in their approaches, are essentially trying to solve the GLT challenge. Benefit from the feature learning, these ensemble learning and data augmentation approaches can also serve as good baselines for the proposed GLT as well. Meanwhile, the proposed IFL is orthogonal to them and can further boost their performances under all three protocols.

\noindent\textbf{Q8: why the proposed GLT is the ``generalized'' version of the LT.} \textbf{A8:} it's because GLT methods would naturally solve the conventional LT, but not vise versa. As shown in Fig.~\ref{fig:8}~(c-d), GLT baselines like augmentation, ensemble, and the proposed IFL automatically solve the class-wise LT. Yet, the majority of class-wise re-balancing LT algorithms cannot tackle the attribute-wise imbalance of GLT very well.




\section{Conclusion}
\label{sec:6}

We introduced a novel Generalized Long-Tailed (GLT) classification problem, completing the previous class-wise long-tailed classification by incorporating the attribute-wise imbalance nature in real-world dataset at scale, which deeply explains the cause of long-tailed performance within each class and the existence of spurious correlations during classification. Experiments on the proposed two benchmarks, ImageNet-GLT and MSCOCO-GLT, demonstrate the limitations of the previous LT algorithms using class-wise adjustment, and the importance of representation learning in GLT. To this end, we present invariant feature learning (IFL) as the first strong baseline for GLT. IFL adopts a metric loss to encourage the learning of invariant features across environments with diverse attribute distributions. As an improved feature backbone, IFL is orthogonal to most of the previous LT algorithms. After plugging IFL into the conventional LT line-up: one-/two-stage re-balance, augmentation, and ensemble, IFL boosts their performances under all protocols in the proposed GLT benchmarks. 

\noindent\textbf{Limitations: } Due to the limited space, we didn't fully explore all kinds of attribute biases in this paper, \eg, benchmarks like BREEDS~\cite{santurkar2020breeds}, MetaShift~\cite{liang2022metashift} or FMoW-WILDS~\cite{koh2021wilds} all have their unique attribute types. In the future, we are going to extend GLT to more diverse tasks and attribute settings.



\subsubsection{Acknowledgements:} This research is supported by the National Research Foundation, Singapore under its AI Singapore Programme (AISG Award No: AISG2-RP-2021-022) and Alibaba-NTU Singapore Joint Research Institute (JRI). We also feel grateful to the computational resources provided by Damo Academy.

\clearpage

\appendix

\section{Appendix} 
This supplementary material includes: 1) implementation details of the proposed IFL and the baseline methods; 2) a detailed analysis of our problem formulation; 3) more evidences of the re-sampling strategy in the proposed environment construction; 4) more details of the dataset construction for the proposed ImageNet-GLT and MSCOCO-GLT benchmarks; 5) more experimental results of the proposed Invariant Feature Learning framework.

\begin{algorithm}[H]
        \caption{The proposed IFL algorithm}
        \begin{algorithmic}
          \STATE \textbf{Input:} the original training set $\{(x,y)\}$
          \STATE \textbf{Initialize:} backbone $f(\cdot;\theta)$, classifier $g(\cdot;w)$
          \FOR {N warm-up epochs}
            \STATE $//$ optimizing the model from cross-entropy classification loss
            \STATE $\theta, w \in \arg\min_{\theta, w} L_{cls}(f(x;\theta), y;w)$
          \ENDFOR
          \STATE \textbf{Initialize:} class centers $\{C_y\}$ for all classes
          \REPEAT
          \STATE $\{(x^{e_1},y^{e_1})\}$, $\{(x^{e_2},y^{e_2})\}$ = EnvConstruct($\{(x,y)\}$, $\theta$, $w$)
          \FOR {M epochs}
             \STATE $//$ learning from $L_{cls}$ and $L_{IFL}$
             \STATE $\theta, w \in \arg\min_{\theta, w} \sum_{e\in \mathcal{E}} \sum_{i\in e} (L_{cls} + \alpha \cdot L_{IFL})$
             \STATE $//$ update class center
             \STATE $\{C_y\} \leftarrow$ MovingAverage($\{C_y\}, \{(f(x^{e_1};\theta), y^{e_1})\}, \{ (f(x^{e_2};\theta), y^{e_2}) \}$)
          \ENDFOR
          \UNTIL{Converge}
          \STATE \textbf{Output: } backbone $f(\cdot;\theta)$, classifier $g(\cdot;w)$
        \end{algorithmic}
        \label{apx_alg:1}
\end{algorithm}

\section{Implementation Details} 

For the proposed IFL, we first trained the model using the cross-entropy loss for 60 epochs, then we started to construct and update environments every 20 epochs. This is because the early epochs are learning generalized features~\cite{prechelt1998early,zhang2021understanding}, so constructing environments in early epochs won't benefit the robustness of the model too much. The trade-off parameter $\alpha$ between the classification loss and IFL metric loss is initialized as 0.0 and changed to 0.001 and 0.005 along with the environment construction. Besides, following the center loss~\cite{wen2016centerloss}, the mean feature is accumulated by an SGD optimizer with the fixed learning rate of 0.5. The pseudo code of the overall IFL algorithm is given in Algorithm~\ref{apx_alg:1}. 

For fair comparisons, we re-implemented all the investigated algorithms into the same GLT codebase that is publicly available on Github: \url{https://github.com/KaihuaTang/Generalized-Long-Tailed-Benchmarks.pytorch}. By default, all images are resized to $112 \times 112$ using Random Resized Crop and Random Horizontal Flip during training. We used ResNeXt-50~\cite{xie2017aggregated} as our backbone for all methods except for BBN~\cite{zhou2019bbn}, RIDE~\cite{wang2020long} and TADE~\cite{zhang2021test}. All models were trained by the SGD optimizer with the batch size 256. The initial learning rate is 0.1 and the default learning rate decay strategy is Cosine Annealing scheduler~\cite{loshchilov2016sgdr} except for BBN~\cite{zhou2019bbn}, LDAM~\cite{cao2019ldam} and RIDE~\cite{wang2020long} that adopted the multi-step scheduler based on original settings in the corresponding papers. For all one-stage methods, results were reported as the performance of the model at epoch 100. For two-stage cRT~\cite{kang2019decoupling} and LWS~\cite{kang2019decoupling}, additional 10 epochs were required to fine-tune a balanced classifier.

\section{Problem Formulation}
In Section~3 of the original paper, we formulate the classification model $p(Y|X)$ as $p(Y|z_c,z_a)$ based on a common assumption~\cite{mirza2014conditional} that any object image $X$ equals to a set of underlying class-specific components $z_c$ and varying attributes $z_a$, \ie, $X=(z_c,z_a)$, where $(z_c,z_a)$ can fully describe the entire the image $X$. Regarding the classification task, a robust feature should only be extracted from the underlying $z_c$, leaving non-robust $z_a$ out of the visual feature. That is to say, an ideal feature backbone $f(\cdot)$ should only respond to class-specific $z_c$: $\mathbf{z}=f(X)=h(z_c)$, where there exists a mapping function $h(\cdot)$ between the extracted feature $\mathbf{z}$ and the underlying robust class-specific components $z_c$. So, we can further convert the $p(Y|X)$ into the following formula using the Bayes theorem~\cite{stone2013bayes}:
\begin{equation}
\begin{split}
    p(Y=k|X=x)&=p(Y=k|z_c, z_a) \\
              &=\frac{p(z_c,z_a|Y=k)}{p(z_c,z_a)} \cdot p(Y=k) \\
              &=\frac{p(z_c|Y=k)}{p(z_c)} \cdot \underbrace{\frac{p(z_a|Y=k,z_c)}{p(z_a|z_c)}}_{attribute\;bias} \cdot \underbrace{p(Y=k)}_{class\;bias},
\end{split}
    \label{apx_eq:1}
\end{equation}
where $\frac{p(z_c|Y=k)}{p(z_c)}$ is a robust indicator of the class; $\frac{p(z_a|Y=k,z_c)}{p(z_a|z_c)}$ is a bias term introduced by the existence of imbalanced attributes; $p(Y=k)$ is a class bias reflecting the class distribution of the training data.

To better understand this formulation, we will provide a more detailed analysis, together with some interesting findings that can be derived from Eq.~\eqref{apx_eq:1}, in the following sub-sections.

\subsection{Details about class-specific Components}

Since the underlying class-specific $z_c$ equals to the existence of multiple independent components, the formal definition of $\frac{p(z_c|Y=k)}{p(z_c)}$ in Eq.~\eqref{apx_eq:1} is as follows:
\begin{equation}
    \frac{p(z_c|Y=k)}{p(z_c)} = \prod_{c_i \in exist} \frac{p(z_{c_i}=1|Y=k)}{p(z_{c_i}=1)} \cdot \prod_{c_j \in non-exist}  \frac{p(z_{c_j}=0|Y=k)}{p(z_{c_j}=0)},
    \label{apx_eq:2}
\end{equation}
where the underlying class-specific components $z_c$ can be regarded as a $0/1$ vector; $\{c_i\}$ are the existed components on the object; $\{c_j\}$ are the non-existed components on the object. An object $X$ belongs to class $Y=k$, if and only if \textbf{I)} all the essential components of class $Y=k$ exist, \eg, a human has to contain the existence of both $p(z_{head\;on\;object}=1|Y=human)=1$ and $p(z_{body\;on\;object}=1|Y=human)=1$, \textbf{II)} and the other irrelevant components don't exist, \eg, $p(z_{tail\;on\;object}=0|Y=human)=1$ and $p(z_{horn\;on\;object}=0|Y=human)=1$. 

Note that the above formulation may raise two questions regarding the class-specific $z_c$: 1) the hierarchy of classification and 2) the partial occlusion.

\textbf{The Hierarchy of Classification:} One common question about the above Eq.~\eqref{apx_eq:2} would be what if an object with all the essential components of a class also contains other irrelevant components, \eg, an object with both human head and human body also has 4 horse limbs\footnote{Centaur: a creature from Greek mythology that has both human upper body and horse lower body~\cite{centaur}, which can be regarded as a sub-species of human.}. This problem is raised from the hierarchy of classification. If an object from class $Y=k$ contains one or more unnecessary components, it usually means the object is actually from a sub-species of class $Y=k$, which is belong to $Y=k$ but with more fine-grained descriptions. In a valid multi-class classification dataset, a species and its sub-species won't simultaneously exist in the prediction vocabulary. Otherwise, it becomes a multi-label classification task. However, multiple sub-species from a common hidden hyper-species can be co-existed in one dataset, \eg, ``shetland sheepdog'' and ``pug-dog'' in ImageNet~\cite{russakovsky2015imagenet} are both from a hyper-species ``dog''. In this case, although these sub-species may have close templates, the class-specific components $z_c$ still satisfy the property of intra-class invariance. In fact, visualizations from the previous research~\cite{tang2020long} found that the cross-entropy loss will automatically learn discriminative components instead of some common components in this case, \eg, only the teeth of ``warthog'' will be learned as class-specific features, rather than the entire body, to increase the power of discrimination within the dataset.

\textbf{The Partial Occlusion:} Another question is about what if there are partial occlusions on the images. Does this break the intra-class invariance of class-specific components $z_c$? The answer is still NO. The recent study of Masked Auto-Encoders~\cite{MaskedAutoencoders2021} proves that deep-learning models can reconstruct the entire image from mere 25\% patches due to the redundant visual information. However, what if the strong occlusion hurts the class-specific component $z_c$ by removing the entire component? During training, these images shouldn't exist in the training data in the first place, as they can not be labeled. During prediction, the corresponding image is no longer predictable and supposed to be detected as an outlier~\cite{abdar2021review}, which is beyond the scope of this paper. Therefore, if there are just some slight partial occlusions, models would still be capable of obtaining the original $z_c$, as we can naturally imagine the entire object from the redundant visual information.

\subsection{Details about the attributes components}
When we look at Eq.~\eqref{apx_eq:1}, we may wonder if there exists certain attribute $z_{a_j}$ that is \textbf{class-independent}, \ie, $p(z_{a_j}|Y=k,z_c) = p(z_{a_j}|z_c)$. The answer is YES. So we can have a formal definition of $\frac{p(z_a|Y=k,z_c)}{p(z_a|z_c)}$ in Eq.~\eqref{apx_eq:1} as follows:
\begin{equation}
    \frac{p(z_a|Y=k,z_c)}{p(z_a|z_c)} = \prod_{a_i \in d-Y} \underbrace{\frac{p(z_{a_i}|Y=k,z_c)}{p(z_{a_i}|z_c)}}_{biased\;attributes} \cdot \prod_{a_j \in ind-Y} \underbrace{\frac{p(z_{a_j}|Y=k,z_c)}{p(z_{a_j}|z_c)}}_{benign\;attributes},
    \label{apx_eq:3}
\end{equation}
where $d-Y$ and $ind-Y$ stand for class-dependent and class-independent, respectively; $\{a_i\}$ are class-dependent attributes; $\{a_j\}$ are class-independent attributes. It's easy to notice that all benign attributes have $p(z_{a_j}|Y=k,z_c) = p(z_{a_j}|z_c)$, so the $\prod_{a_j \in ind-Y} \frac{p(z_{a_j}|Y=k,z_c)}{p(z_{a_j}|z_c)}$ always equals to $1$ and won't introduce any biases. In the original paper, we consider all attributes as a whole, \ie, a vector $z_a$, so we didn't differentiate the biased attributes and benign attributes. Their differences are explained in the following.

\textbf{Biased Attributes:} Nearly all semantic attributes are biased. For example, considering a class-specific component $z_{c_{fur}}$ as ``fur'' and $z_{a_{color}}$ is an attribute describing its ``color'', different animals (Class Y) may have different distributions for the color of fur. In summary, the biased attributes $\{a_i \in d-Y\}$ are either dependent on a class-specific component, \eg, the ``color'' or ``texture'' for a specific $z_{c_i}$, or dependent on multiple components at the same time, \eg, the ``posture'' of a human that depends on both head, body and four limbs. Those biased attributes not only cause the long-tailed prediction confidence within each class, but also create spurious correlations between a biased attribute $z_{a_i}$ and a specific class $Y=k$. 

\textbf{Benign Attributes:} Since we consider all attributes as a whole in the original paper, \ie, an underlying vector $z_a$, we don't differentiate benign attributes from the entire $z_a$. However, if we look as Eq.~\eqref{apx_eq:3}, an attribute $a_j \in ind-Y$ is totally benign as its distribution is independent of $Y=k$, \ie, $p(z_{a_j}|Y=k,z_c) = p(z_{a_j}|z_c)$, making $\frac{p(z_{a_j}|Y=k,z_c)}{p(z_{a_j}|z_c)} = 1$ become an unbiased term. In fact, \textbf{it fundamentally explains the effectiveness of the intuitive Data Augmentation method}, as the data augmentation introduces additional images (at no cost) with benign attributes that are independent of their classes. There are two types of benign attributes regarding different types of data augmentation: \textbf{I)} the image-level attributes, that are independent of both $z_c$ and $Y$, are commonly used as a pre-processing augmentation in the data loader, as they can be easily generated from any given image, \eg, the rotation (Rotation), position (Crop), size (Resize), and color (ColorJitter) for the image; \textbf{II)} the component-related attributes, that are only independent of $Y$ but still depend on $z_c$, are data augmentation at the data collection stage, \eg, the viewpoint, as they are variations caused by the relative position between the camera and a specific component instead of the object class. These benign attributes can increase both the volume and the diversity of the classification dataset without any cost, which explains why the data augmentation method like RandAug~\cite{cubuk2020randaugment} are so effective in our experiments.

\section{Environment Construction} 

\begin{figure}[t]
   \begin{minipage}[b]{1.0\linewidth}
   \centerline{\includegraphics[width=100mm]{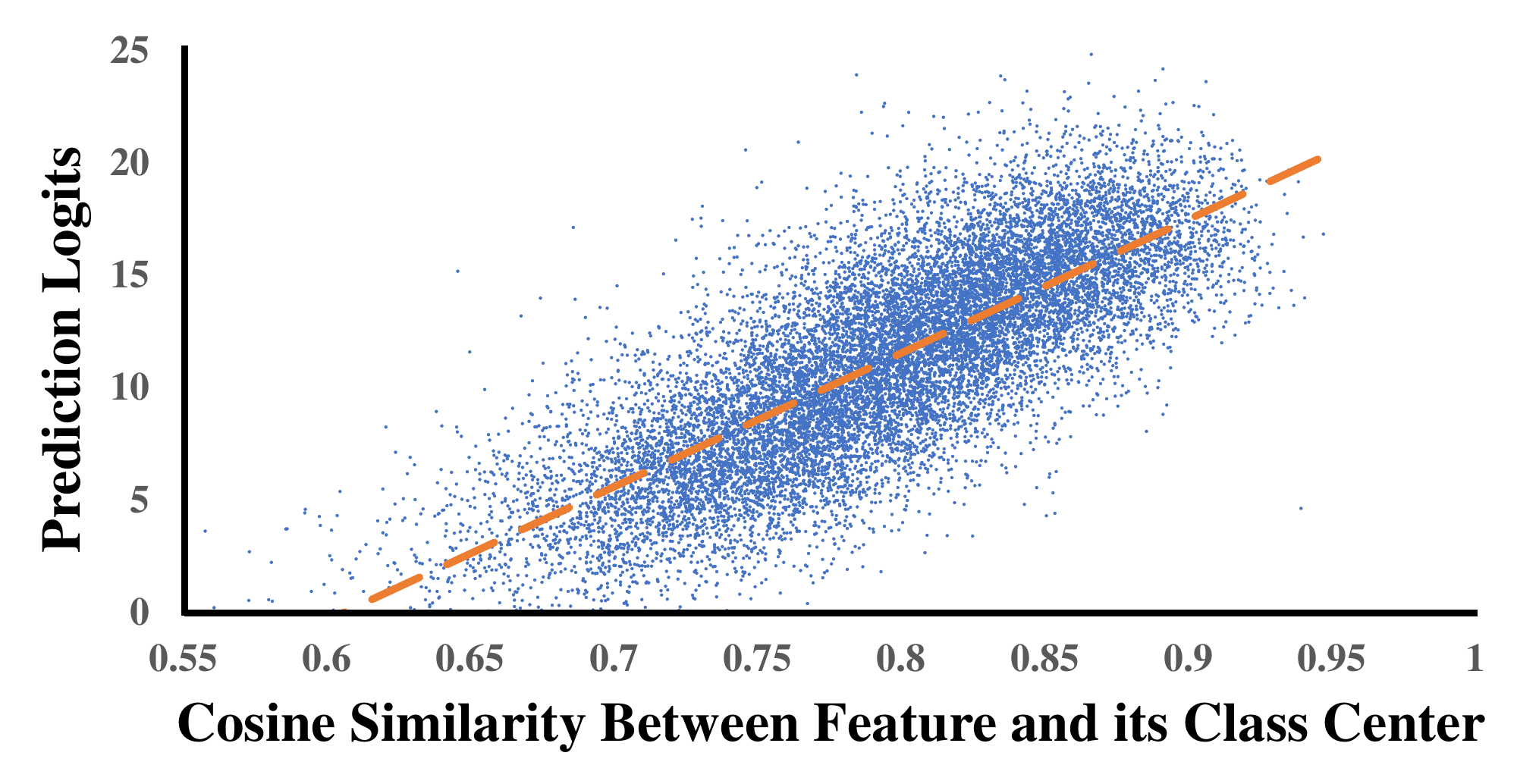}}
   \end{minipage}
   \caption{The relationship between the prediction logits and the cosine similarity between image features and its class center in ImageNet-GLT. The direct proportion between them demonstrates why the prediction confidence, \ie, $p(Y=k|X)$, can be used to probe the intra-class variation, \ie, $\frac{p(z_a|Y=k,z_c)}{p(z_a|z_c)}$.}
   \label{fig:apx1} 
\end{figure}

\subsection{Motivation}
Since we don't impose the disentanglement assumption that perfect feature vectors $\mathbf{z}=[z_c;z_a]$ with separated $z_c$ and $z_a$ can be learned, we cannot easily eliminate $z_a$ through feature selection. Therefore, we can only implicitly prevent features from associating with $z_a$ during the backbone optimization. To achieve this, we use the prediction logits of images to sample diverse distributions of $\frac{p(z_a|Y=k,z_c)}{p(z_a|z_c)}$ based on the formulation Eq.~\eqref{apx_eq:1}, as $p(Y=k|X) \propto \frac{p(z_a|Y=k,z_c)}{p(z_a|z_c)}$ within a given class $Y=k$. If the feature backbone relies on the non-robust $z_a$, the diverse distributions of $\frac{p(z_a|Y=k,z_c)}{p(z_a|z_c)}$ will make the model have unstable class centers across different environments. On the contrary, if the feature backbone relies more on the robust and invariant components $z_c$, the class centers should be more consistent against the environment change, which motivates the design of the proposed IFL. 

\subsection{Re-sampling Strategy}
The overall environment construction pipeline can be explained as follows: since $\frac{p(z_a|Y=k,z_c)}{p(z_a|z_c)}$ is only directly proportional to $p(Y=k|X)$ within a given class $Y=k$, the re-sampling strategy is conducted within each class independently. Then, for each environment, we merge the resultant subsets from each class that use the same re-sampling strategy, so the diverse environments are thus corresponding to different re-sampling strategy, \ie, different $\frac{p(z_a|Y=k,z_c)}{p(z_a|z_c)}$. 

In this sub-section, we will provide additional evidences to support the re-sampling strategy in our  environment construction method.
To prove the direct correlation between $p(Y=k|X)$ and $\frac{p(z_a|Y=k,z_c)}{p(z_a|z_c)}$, we visualize the relationship between the prediction logits and the cosine similarity between image features and its class center in Fig.~\ref{fig:apx1}. First of all, $p(z_a|z_c)$ can be regarded as a fixed layout in the given dataset and $\frac{p(z_c|Y=k)}{p(z_c)}$ is invariant within a class, therefore, the intra-class variation is mainly caused by $p(z_a|Y=k,z_c)$ in a given class and dataset. Since the attribute $z_a$ is long-tailed, the prediction logits should also exhibit long-tailed distribution, \ie, most of the samples with few head attributes have relatively larger logits while few samples with most of the rare attributes have lower logits. The visualized Fig.~\ref{fig:apx1} proved this assumption. As the class center is dominated by the head attributes due to their large population, the cosine similarity between the feature of an image and its class center actually indicates the rarity of the attribute vector $z_a$ for this image. The rarer the attributes are, the lower the prediction confidence is.

In summary, Fig.~\ref{fig:apx1} demonstrates the reason why we can use $p(Y=k|X)$ to probe the $\frac{p(z_a|Y=k,z_c)}{p(z_a|z_c)}$ and thus construct diverse environments and how bias the intra-class distribution is (caused by long-tailed attributes).

\begin{figure}
   \begin{minipage}[b]{1.0\linewidth}
   \centerline{\includegraphics[width=150mm]{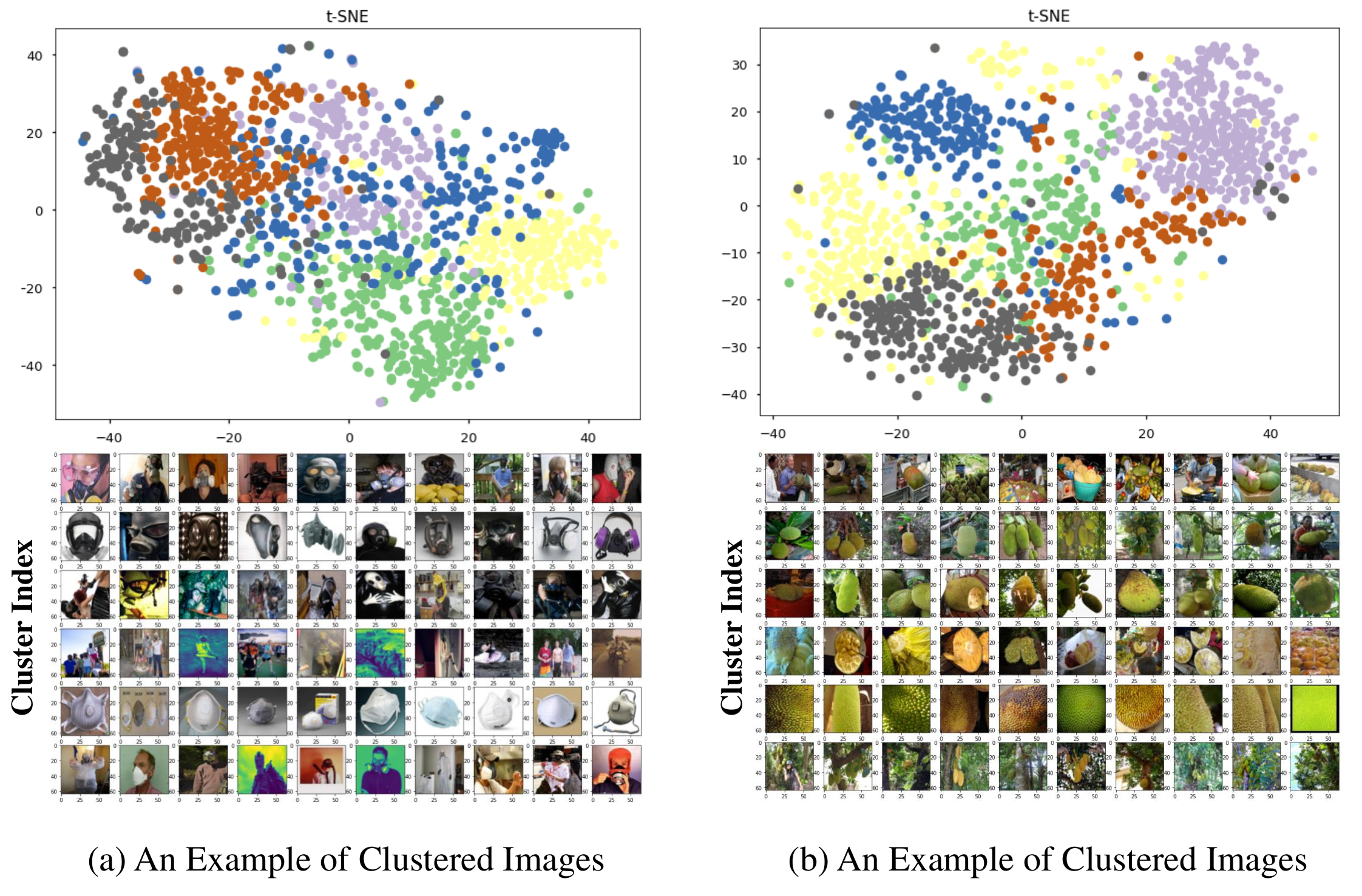}}
   \end{minipage}
   \caption{Examples of feature clusters using KMeans (through t-SNE~\cite{hinton2002stochastic}) and the corresponding visualized images for each cluster in the original ImageNet~\cite{russakovsky2015imagenet}.}
   \label{fig:apx2} 
\end{figure}

\section{Dataset Construction}

\subsection{ImageNet-GLT}

The proposed ImageNet-GLT benchmark is a long-tailed subset generated from the original ImageNet~\cite{russakovsky2015imagenet} dataset. For Train-GLT, Train-CBL, and Test-CBL splits, we can directly follow the previous ImageNet-LT benchmark~\cite{liu2019large} to construct these splits. The tricky part is the Test-GBL that is used to evaluate the Attribute-wise Long Tail (ALT) protocol and Generalized Long Tail (GLT) protocol. To create the attribute-wise balanced and class-wise balanced evaluation environment (Test-GBL), we used the index of several feature clusters of each category as the pretext attribute labels.

\begin{figure}
   \begin{minipage}[b]{1.0\linewidth}
   \centerline{\includegraphics[width=150mm]{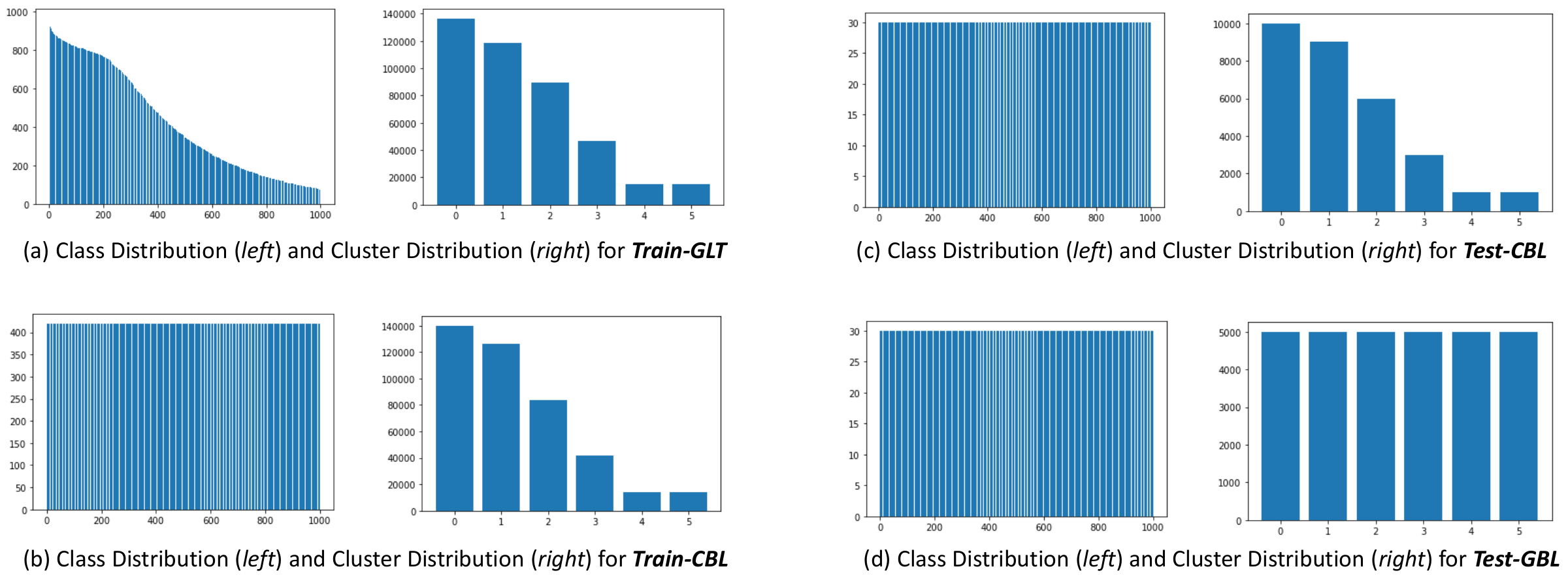}}
   \end{minipage}
   \caption{Class distributions and cluster distributions for each split of ImageNet-GLT benchmark. Note that clusters may represent different attribute layouts in each classes, so ImageNet-GLT actually has $6 \times 1000$ pretext attributes rather than 6, \ie, each column in the cluster distribution stands for 1000 pretext attributes having the same frequency.}
   \label{fig:apx5} 
\end{figure}

\begin{figure}
   \begin{minipage}[b]{1.0\linewidth}
   \centerline{\includegraphics[width=150mm]{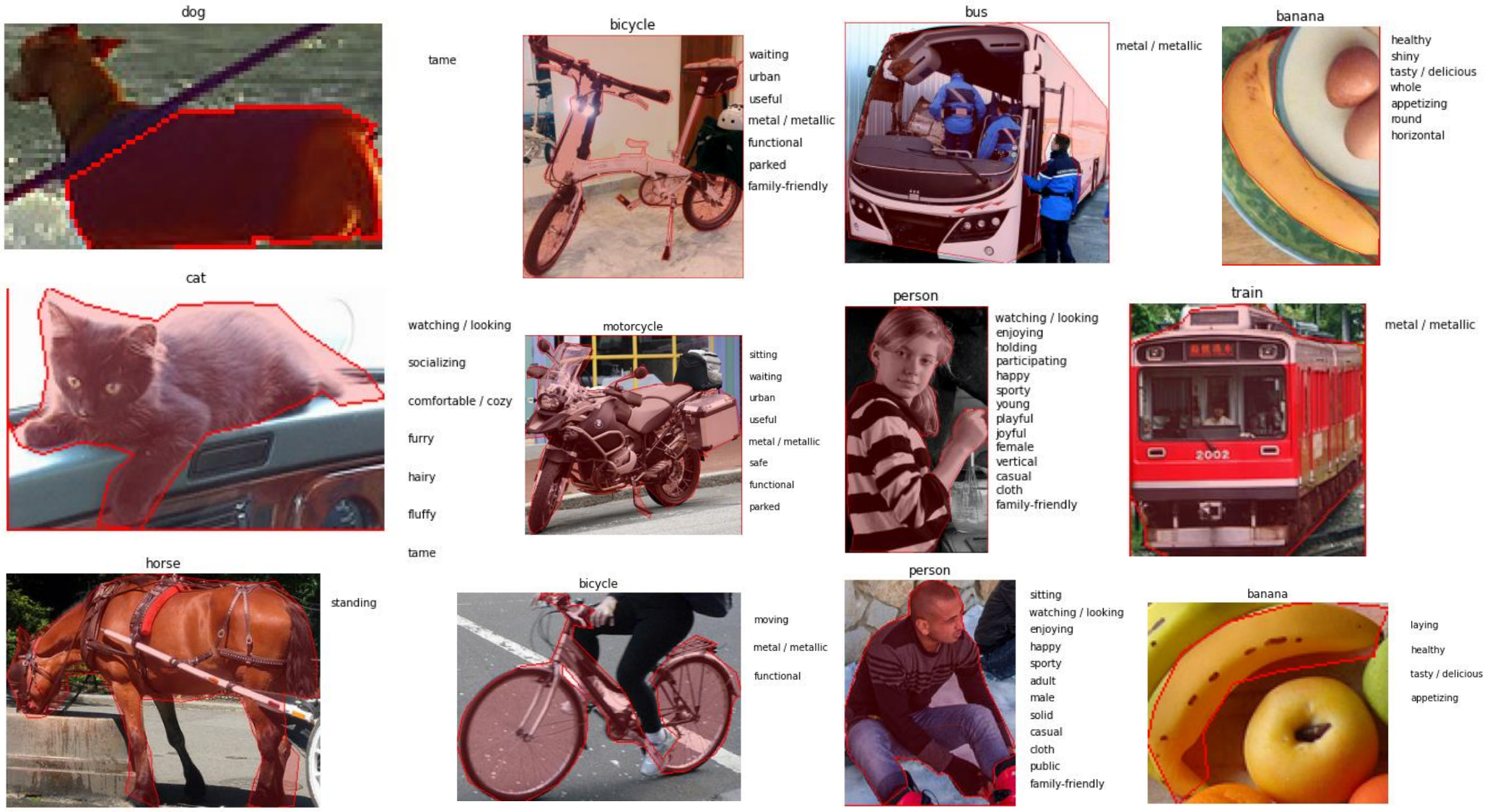}}
   \end{minipage}
   \caption{Examples of objects from MSCOCO-Attribute~\cite{patterson2016coco} dataset. Note that the object attribute won't be released in the proposed MSCOCO-GLT}
   \label{fig:axp3} 
\end{figure}

To begin with, we applied a pre-trained ResNet-50~\cite{he2016deep} backbone provided by PyTorch~\cite{NEURIPS2019_9015} to extract a 2048-dimensional feature vector for all images in the ImageNet dataset. Then, we ran the KMeans algorithm to generate 6 clusters for each category. As we can see from Figure~\ref{fig:apx2}, the corresponding visualized images demonstrate that the feature clusters within each category naturally separate images by different types of attributes, \eg, materials, backgrounds, \etc.

Afterwards, we smoothed the attribute distribution of each category by ensuring the frequencies of Top-2, Medium-2, Bottom-2 clusters to be 70\%, 20\%, and 10\% for all categories. Note that the underlying attribute distribution is naturally long-tailed. However, since we only used a limited number of clusters, \ie, 6 clusters, to approximate the numerous realistic attributes, those clusters could be relatively too balanced for some categories, so we force the distribution of pretext attributes to be the same for all categories.

We also list all the class distributions and cluster distributions for all four split, Train-GLT, Train-CBL, Test-CBL, Test-GBL, of ImageNet-GLT benchmark in Fig.~\ref{fig:apx5}. Note that each column in the cluster distribution stands for 1000 pretext attributes for all 1000 classes that have the same frequency, as clusters may represent different attribute layouts in each classes.


\begin{figure}
   \begin{minipage}[b]{1.0\linewidth}
   \centerline{\includegraphics[width=150mm]{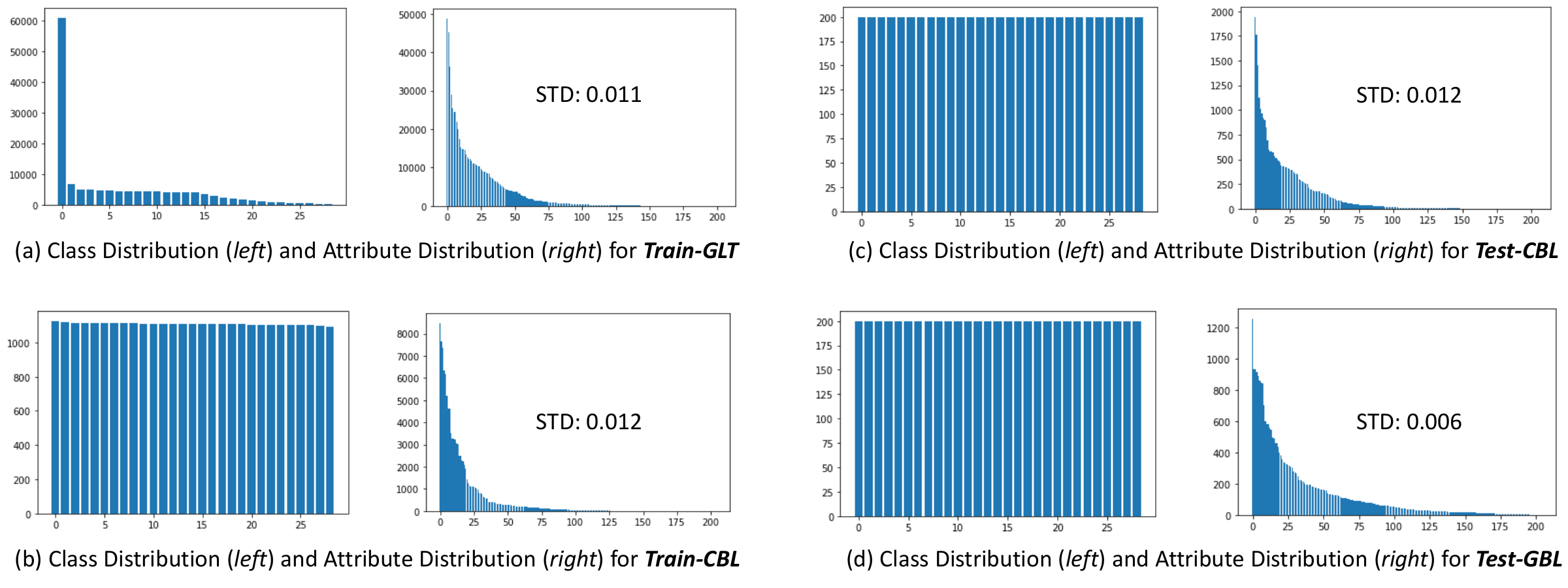}}
   \end{minipage}
   \caption{Class distributions and attribute distributions for each split of MSCOCO-GLT benchmark, where the most frequent category is the ``person''. Although we cannot strictly balance the attribute distribution in Test-GBL split due to the fact that both head attributes and tail attributes can co-occur in one object, the selected Test-GBL has lower standard deviation of attributes than other splits.}
   \label{fig:apx6} 
\end{figure}

\begin{table*}
\centering
\caption{\textbf{Evaluation of GLT Protocol on ImageNet-GLT}: This is a supplementary table of the Table~1 in original paper. Many$_A$, Medium$_A$, Few$_A$ indicate the head (Top-2 Clusters), medium (Medium-2 Clusters), and tail (Bottom-2 Clusters) attributes. The (Top-1) Accuracy and Precision of previous LT algorithms and their IFL variants are reported}
\scalebox{0.75}
{
\begin{tabu}{c| c |[1.5pt]c |c |c |c |c |c |c |c}
\hline
\hline
\multicolumn{2}{c|[1.5pt]}{Setting} & \multicolumn{8}{c}{Generalized Long Tail (GLT) Protocol} \\ 
\hline 
\multicolumn{2}{c|[1.5pt]}{Test Splits} & \multicolumn{2}{c|}{Many$_A$} & \multicolumn{2}{c|}{Medium$_A$} & \multicolumn{2}{c|}{Few$_A$} & \multicolumn{2}{c}{Overall} \\ 
\hline 
\multicolumn{2}{c|[1.5pt]}{Evaluation Metric} & Accuracy & Precision & Accuracy & Precision & Accuracy & Precision & Accuracy & Precision \\
\hline

\multirow{12}{*}{{\rotatebox{90}{\small{\textbf{Re-balance}}}}} 

&Baseline & 47.36 & 51.84 & 33.45 & 36.05 & 23.44 & 25.66 & 34.75 & 40.65 \\

&cRT~\cite{kang2019decoupling} & 51.21 & 50.48 & 36.25 & 35.03 & 25.24 & 24.62 & 37.57 & 37.51 \\

&LWS~\cite{kang2019decoupling} & 51.95 & 51.14 & 36.77 & 35.60 & 25.09 & 24.70 & 37.94 & 38.01 \\

&Deconfound-TDE~\cite{tang2020long} & 51.03 & 49.52 & 36.11 & 34.25 & 25.53 & 24.47 & 37.56 & 37.00 \\

&BLSoftmax~\cite{ren2020balanced} & 50.58 & 50.21 & 35.74 & 35.17 & 24.96 & 25.46 & 37.09 & 38.08 \\

&Logit-Adj~\cite{menon2020long} & 52.25 & 51.08 & 36.12 & 34.92 & 25.04 & 24.71 & 37.80 & 37.56 \\

&BBN~\cite{zhou2019bbn} & 51.92 & 53.17 & 36.24 & 36.26 & 25.56 & 26.07 & 37.91 & 41.77 \\

&LDAM~\cite{cao2019ldam} & 52.33 & 51.91 & 37.09 & 36.07 & 26.19 & 25.56 & 38.54 & 39.08 \\

&(ours) Baseline + IFL & 51.16 & \textbf{56.64} & 36.50 & 39.88 & 26.21 & 29.00 & 37.96 & \textbf{44.47} \\

&(ours) cRT + IFL & 53.53 & 54.93 & 38.15 & 38.21 & 27.13 & 27.66 & 39.60 & 41.65 \\

&(ours) LWS + IFL & 53.31 & 55.42 & 38.21 & 39.16 & 27.40 & 28.46 & 39.64 & 42.45 \\

&(ours) BLSoftmax + IFL & 53.68 & 55.35 & 39.05 & \textbf{40.67} & 27.49 & 29.09 & 40.08 & 43.48 \\

&(ours) Logit-Adj + IFL & \textbf{54.90} & 55.19 & \textbf{39.09} & 39.93 & \textbf{27.57} & \textbf{29.27} & \textbf{40.52} & 42.28 \\

\tabucline[1.5pt]{-}

\multirow{4}{*}{{\rotatebox{90}{\small{\textbf{Augment}}}}} 

&Mixup~\cite{zhang2018mixup} & 43.46 & 47.05 & 30.23 & 30.63 & 20.96 & 22.73 & 31.55 & 37.44 \\

&RandAug~\cite{cubuk2020randaugment} & 51.63 & 55.21 & 36.96 & 39.95 & 26.14 & 27.84 & 38.24 & 44.74 \\

&(ours) Mixup + IFL & 57.10 & 61.66 & 41.46 & 45.04 & 30.43 & 33.42 & 43.00 & 49.25 \\

&(ours) RandAug + IFL & \textbf{58.67} & \textbf{62.42} & \textbf{44.06} & \textbf{47.36} & \textbf{31.97} & \textbf{34.67} & \textbf{44.90} & \textbf{50.47} \\

\tabucline[1.5pt]{-}

\multirow{4}{*}{{\rotatebox{90}{\small{\textbf{Ensemble}}}}} 

&TADE~\cite{zhang2021test} & 55.52 & 56.35 & 40.86 & 41.23 & 28.85 & 30.01 & 41.75 & 44.15 \\

&RIDE~\cite{wang2020long} & 57.49 & 56.24 & 41.57 & 39.96 & 29.94 & 29.18 & 43.00 & 43.32 \\

&(ours) TADE + IFL & 56.99 & 57.44 & 42.55 & 42.39 & 30.86 & 31.55 & 43.47 & 45.17 \\

&(ours) RIDE + IFL & \textbf{59.24} & \textbf{59.87} & \textbf{44.98} & \textbf{44.80} & \textbf{32.69} & \textbf{32.55} & \textbf{45.64} & \textbf{47.14} \\

\hline
\hline
\end{tabu}
}
\label{tab:apx1}
\end{table*}

\subsection{MSCOCO-GLT}
The proposed MSCOCO-GLT benchmark is a long-tailed subset generated from the MSCOCO-Attribute~\cite{patterson2016coco,lin2014microsoft}, where we cropped each object as individual images. Fig.~\ref{fig:axp3} shows several examples of objects from MSCOCO-Attribute~\cite{patterson2016coco} dataset. Note that the object attribute won't be released in the proposed MSCOCO-GLT.

Different from the ImageNet-GLT, we adopt real attribute annotations to construct the MSCOCO-GLT. However, as we can see from Fig.~\ref{fig:axp3}, a single object can be labelled with multiple attributes, causing the tail attribute co-occurring with the head in one image. Therefore, it's impossible to strictly balance the attribute distribution in the Test-GBL, so we proposed the Algorithm~\ref{alg:gbl} to generate the Test-GBL for MSCOCO-GLT. Specifically, we need to iteratively look for samples to minimize the standard deviation of the overall attribute distribution within each category, so the output subset is the most balanced subset in terms of the attribute distribution in the entire dataset.

As to the Train-GLT, Train-CBL, and Test-CBL, we can use the same strategy as the previous sub-section to directly sample from each category. Note that the long-tailed distribution in the MSCOCO-GLT is more severe than ImageNet-GLT, as it only has 29 classes but the single class ``person'' possess over 40\% of the training data. Therefore, the class-wise balanced Train-CBL has much smaller size of data. 

We also list all the class distributions and attribute distributions for all four split, Train-GLT, Train-CBL, Test-CBL, Test-GBL, of MSCOCO-GLT benchmark in Fig.~\ref{fig:apx6}. Here we only show the Test-GBL for CLT and GLT protocols. Test-GBL for ALT protocols has small size due to the cost of class-wise data re-balance for Train-CBL, but the class and attribute distributions are still the same as Fig.~\ref{fig:apx6}~(d). It also worth noting that although the attribute distribution is not strictly balanced in Fig.~\ref{fig:apx6}~(d), it has the lowest attribute STD, making the Test-GBL more balanced in attributes than other splits.

\begin{algorithm}[H]
\caption{Generating Test-GBL for MSCOCO-GLT}
\label{alg:gbl}
\begin{algorithmic}
 \STATE \textbf{Input}: $\{(x,y,z)\}$ is a set of object $x$ with corresponding label $y$ and attribute $z$
 \STATE \textbf{Parameter}: $N$ is the number of selected images for each category
 \STATE \textbf{Initialize}: \textbf{Test-GBL} = $\{\}$
 \FOR{\textit{i} in \textbf{Y}}   
    \STATE //Y is the set of category
        \STATE $count = 0$
        \STATE $z_{dist} = [0,0,0,...]$
        \WHILE{$count < N$}
            \STATE $Min_{STD}= MAX$
            \STATE $Cancidate = None$
             \FOR{\texttt{$(x,y,z)$ in $\{(x,y,z)\}$}} 
                \IF{$y=i$}
                    \STATE $temp_{dist} = z_{dist} + z$
                    \STATE $temp_{dist} = Norm(temp_{dist})$
                    \STATE $temp_{STD} = STD(temp_{dist})$
                    \IF{$temp_{STD} < Min_{STD}$}
                        \STATE $ Min_{STD} = temp_{STD}$
                        \STATE $Cancidate = (x,y)$
                        \STATE $z_{candidate} = z$
                    \ENDIF
                \ENDIF
             \ENDFOR
        \STATE \textbf{Test-GBL}.append($Candidate$)
        \STATE $z_{dist} = z_{dist} + z_{candidate}$
        \STATE $count += 1$
        \ENDWHILE
\ENDFOR
\STATE \textbf{Output}: \textbf{Test-GBL}
\end{algorithmic}
\end{algorithm}

\begin{table}[t]
\centering
\caption{\textbf{Evaluation of CLT and GLT Protocols on MSCOCO-GLT}: This is a supplementary table of the Table~3 in original paper. (Top-1) Accuracy $\vert$ Precision of previous LT algorithms and their variants equipped with the proposed IFL are reported. All methods are re-implemented under the same codebase with ResNext-50 backbone to ensure fair comparisons}
\scalebox{0.58}
{
\begin{tabu}{c| c |[1.5pt]c |c |c |c |[1.5pt]c |c |c |c }
\hline
\hline
\multicolumn{2}{c|[1.5pt]}{Methods} & \multicolumn{4}{c|[1.5pt]}{\textbf{Class-wise Long Tail (CLT) Protocol}} & \multicolumn{4}{c}{\textbf{Generalized Long Tail (GLT) Protocol}} \\ 
\hline 
\multicolumn{2}{c|[1.5pt]}{\textbf{$<$ Accuracy $\vert$ Precision $>$}} & Many$_C$ & Medium$_C$ & Few$_C$ & Overall & Many$_C$ & Medium$_C$ & Few$_C$ & Overall\\ 
\hline 

\multirow{13}{*}{{\rotatebox{90}{\textbf{Re-balance}}}} 

& Baseline & 80.86 $\vert$ 71.21 & 74.79 $\vert$ 76.35 & 51.83 $\vert$ 87.01 & 72.34 $\vert$ 76.61 & 74.41 $\vert$ 65.54 & 66.58 $\vert$ 69.81 & 38.75 $\vert$ 81.08 & 63.79 $\vert$ 70.52 \\

& cRT~\cite{kang2019decoupling} & 80.18 $\vert$ 71.18 & 75.71 $\vert$ 75.43 & 57.50 $\vert$ 85.19 & 73.64 $\vert$ 75.84 & 73.00 $\vert$ 64.34 & 67.42 $\vert$ 68.00 & 44.00 $\vert$ 76.32 & 64.69 $\vert$ 68.33 \\

& LWS~\cite{kang2019decoupling} & 80.50 $\vert$ 69.85 & 74.46 $\vert$ 75.21 & 54.42 $\vert$ 87.20 & 72.60 $\vert$ 75.66 & 73.55 $\vert$ 63.57 & 66.08 $\vert$ 67.57 & 40.42 $\vert$ 80.87 & 63.60 $\vert$ 68.81 \\

& Deconfound-TDE~\cite{tang2020long} & 78.36 $\vert$ 72.64 & 77.96 $\vert$ 73.01 & 57.08 $\vert$ 82.83 &  73.79 $\vert$ 74.90 & 72.14 $\vert$ 65.99 & 71.04 $\vert$ 66.68 & 45.00 $\vert$ 75.29 & 66.07 $\vert$ 68.20 \\

& BLSoftmax~\cite{ren2020balanced} & 80.41 $\vert$ 73.53 & 77.38 $\vert$ 70.53 & 48.92 $\vert$ 87.86 & 72.64 $\vert$ 75.25 & 72.32 $\vert$ 65.88 & 70.63 $\vert$ 64.21 & 35.83 $\vert$ 82.30 & 64.07 $\vert$ 68.59 \\

& Logit-Adj~\cite{menon2020long} & 80.36 $\vert$ 74.03 & 78.00 $\vert$ 77.16 & 61.58 $\vert$ 81.54 & 75.50 $\vert$ 76.88 & 73.64 $\vert$ 66.14 & 68.75 $\vert$ 69.16 & 47.33 $\vert$ 70.79 & 66.17 $\vert$ 68.35 \\

& BBN~\cite{zhou2019bbn} & 82.32 $\vert$ 70.35 & 78.33 $\vert$ 77.35 & 48.58 $\vert$ \textbf{90.22} & 73.69 $\vert$ 77.35 & 75.82 $\vert$ 62.96 & 69.21 $\vert$ 69.59 & 34.25 $\vert$ 84.70 & 64.48 $\vert$ 70.20 \\

& LDAM~\cite{cao2019ldam} & 82.05 $\vert$ 75.43 & 80.04 $\vert$ 74.36 & 54.75 $\vert$ 88.54 & 75.57 $\vert$ 77.70 & 76.05 $\vert$ \textbf{69.93} & 72.96 $\vert$ 66.85 & 39.75 $\vert$ 79.81 & 67.26 $\vert$ 70.70 \\

& (ours) Baseline + IFL & 82.09 $\vert$ 72.40 & 77.50 $\vert$ 79.29 & 53.67 $\vert$ 90.03 & 74.31 $\vert$ 78.90 & 76.09 $\vert$ 66.53 & 68.08 $\vert$ \textbf{71.80} & 40.00 $\vert$ 83.57 & 65.31 $\vert$ \textbf{72.24} \\

& (ours) cRT + IFL & 82.55 $\vert$ 73.93 & 78.58 $\vert$ 79.34 & 59.83 $\vert$ 88.13 & 76.21 $\vert$ 79.11 & \textbf{76.77} $\vert$ 67.28 & 69.25 $\vert$ 70.25 & 44.08 $\vert$ 80.94 & 66.90 $\vert$ 71.34 \\

& (ours) LWS + IFL & 82.27 $\vert$ 73.69 & 78.29 $\vert$ 79.39 & 59.83 $\vert$ 88.82 & 75.98 $\vert$ \textbf{79.18} & 76.73 $\vert$ 67.30 & 68.58 $\vert$ 70.65 & 43.83 $\vert$ 80.88 & 66.55 $\vert$ 71.49 \\

& (ours) BLSoftmax + IFL & 81.77 $\vert$ 72.79 & 78.25 $\vert$ 74.53 & 49.92 $\vert$ 90.07 & 73.72 $\vert$ 77.08 & 74.73 $\vert$ 66.54 & 69.71 $\vert$ 65.39 & 36.58 $\vert$ \textbf{85.58} & 64.76 $\vert$ 70.00 \\

& (ours) Logit-Adj + IFL & \textbf{82.77} $\vert$ \textbf{75.42} & \textbf{79.25} $\vert$ \textbf{79.61} & \textbf{62.67} $\vert$ 84.78 & \textbf{77.16} $\vert$ 79.09 & 75.59 $\vert$ 68.16 & \textbf{69.83} $\vert$ 70.73 & \textbf{48.17} $\vert$ 72.77 & \textbf{67.53} $\vert$ 70.18 \\

\tabucline[1.5pt]{-}

\multirow{4}{*}{{\rotatebox{90}{\small{\textbf{Augment}}}}} 

& Mixup~\cite{zhang2018mixup} & 81.86 $\vert$ 72.96 & 76.58 $\vert$ 80.00 & 55.50 $\vert$ 86.19 & 74.22 $\vert$ 78.61 & 75.41 $\vert$ 67.13 & 67.13 $\vert$ 71.47 & 39.00 $\vert$ 77.79 & 64.45 $\vert$ 71.13 \\

& RandAug~\cite{cubuk2020randaugment} & 83.27 $\vert$ 74.70 & 80.04 $\vert$ 80.76 & 58.50 $\vert$ 87.62 & 76.81 $\vert$ 79.88 & 77.18 $\vert$ 68.07 & 71.46 $\vert$ 72.68 & 42.83 $\vert$ 81.40 & 67.71 $\vert$ 72.73 \\

& (ours) Mixup + IFL & 84.59 $\vert$ \textbf{75.20} & 80.17 $\vert$ \textbf{83.01} & \textbf{59.42} $\vert$ \textbf{91.38} & 77.55 $\vert$ \textbf{81.78} & 78.73 $\vert$ \textbf{69.93} & \textbf{72.21} $\vert$ \textbf{75.50} & \textbf{43.92} $\vert$ 82.52 & \textbf{68.83} $\vert$ \textbf{74.84} \\

& (ours) RandAug + IFL &  \textbf{84.86} $\vert$ \textbf{75.20} & \textbf{81.42} $\vert$ 82.12 & 57.17 $\vert$ 89.85 & \textbf{77.71} $\vert$ 81.10 & \textbf{78.82} $\vert$ 69.19 & 71.42 $\vert$ 73.88 & 42.08 $\vert$ \textbf{82.93} & 68.16 $\vert$ 73.97 \\

\tabucline[1.5pt]{-}

\multirow{4}{*}{{\rotatebox{90}{\small{\textbf{Ensemble}}}}} 

& TADE~\cite{zhang2021test} & 83.77 $\vert$ 76.29 & 81.46 $\vert$ 75.19 & 51.92 $\vert$ 90.80 & 76.22 $\vert$ 78.84 & 75.68 $\vert$ 68.78 & 73.83 $\vert$ 66.27 & 37.33 $\vert$ 85.58 & 66.98 $\vert$ 71.22 \\

& RIDE~\cite{wang2020long} & 83.82 $\vert$ 77.49 & \textbf{84.00} $\vert$ 77.38 & 56.75 $\vert$ 91.44 & 78.29 $\vert$ 80.33 & \textbf{77.09} $\vert$ \textbf{70.22} & \textbf{74.54} $\vert$ 68.50 & 41.08 $\vert$ 83.23 & 68.59 $\vert$ 72.20 \\

& (ours) TADE + IFL & 83.91 $\vert$ 75.78 & 81.63 $\vert$ 75.74 & 52.83 $\vert$ \textbf{92.14} & 76.53 $\vert$ 79.15 & 76.82 $\vert$ 68.55 & 73.88 $\vert$ 68.13 & 37.08 $\vert$ \textbf{88.10} & 67.38 $\vert$ 72.42 \\

& (ours) RIDE + IFL & \textbf{85.05} $\vert$ \textbf{78.36} & 83.21 $\vert$ \textbf{78.68} & \textbf{58.83} $\vert$ 89.04 & \textbf{78.86} $\vert$ \textbf{80.70} & \textbf{77.09} $\vert$ 69.64 & \textbf{74.54} $\vert$ \textbf{70.75} & \textbf{43.50} $\vert$ 81.59 & \textbf{69.09} $\vert$ \textbf{72.57} \\

\hline
\hline
\end{tabu}
}
\label{tab:apx2}
\end{table}

\section{Experimental Results}

Due to the limited space, we simplified the Table~1 and Table~3 in the original paper by skipping the results of some detailed splits, \eg, Many$_A$, Medium$_A$, Few$_A$ in GLT protocol for Table~1 and  Many$_C$, Medium$_C$, Few$_C$ for Table~3. Therefore, we complete the corresponding parts in this supplementary material. 

As we can see from Table~\ref{tab:apx1}, the detailed Many$_A$, Medium$_A$, Few$_A$ of GLT protocol are quite similar to those in the ALT protocol, so the conclusions drawn from the Table~2 of the original still hold at here. We can notice that the effects caused by the attribute-wise imbalance is different from those caused by class-wise imbalance, as there are no precision-accuracy trade-off in Many$_A$, Medium$_A$, Few$_A$. It further proves that why the previous trends~\cite{zhang2021test} of improving both head and tail categories are actually trying to solve the GLT with attribute-wise imbalance, because extracting more attribute-invariant features can benefit both head and tail categories. 

In Table~\ref{tab:apx2}, we completed the Many$_C$, Medium$_C$, Few$_C$ for both CLT and GLT protocols in MSCOCO-GLT, where Many$_C$ contains Top-1 to Top-11 frequent categories, Medium$_C$ has Top-12 to Top-23 categories, and the rest belongs to Few$_C$. The proposed variants of IFL still outperform the corresponding LT methods in most cases. But we notice that the previous strong GLT baseline RandAug + IFL is now worse than another GLT baseline Mixup + IFL. We believe that this is probably caused by the weird distribution in Fig.~\ref{fig:apx6}~(a), where one single super head class ``person'' contains over 40\% of the entire data, so how to learn better boundaries between ``person'' class and other classes becomes the biggest problem, making the Random Augmentation~\cite{cubuk2020randaugment} less effective than Mixup~\cite{zhang2018mixup}, because the latter can ensure the boundaries having linear transition between classes.

\clearpage
%
%
\bibliographystyle{splncs04}
\bibliography{egbib}
\end{document}